\documentclass[letterpaper, preprint, paper,11pt]{AAS}	

\usepackage{bm}
\usepackage{amsmath}
\usepackage{caption}
\usepackage{subcaption}
\usepackage{graphicx}
\usepackage{comment}
\usepackage{textcomp}
\usepackage{algorithm}
\usepackage{algpseudocode}
\usepackage{nicematrix}
\usepackage{arydshln}
\usepackage{multirow}
\usepackage{textcomp}
\usepackage{url}

\usepackage[colorlinks=true, pdfstartview=FitV, linkcolor=black, citecolor= black, urlcolor= black]{hyperref}
\usepackage{overcite}
\usepackage{footnpag}			      	

\begin{document}
\raggedbottom

\title{A Scalable Tabletop Satellite Automation Testbed: Design and Experiments}

\author{Deep Parikh \thanks{Graduate Research Assistant,  Land, Air and Space Robotics (LASR) Laboratory, Aerospace Engineering},
Ali Hasnain Khowaja \footnotemark[1],
Nathan Long \footnotemark[1],
Ian Down \footnotemark[1],
James McElreath \footnotemark[1],
Aniket Bire \footnotemark[1],
Manoranjan Majji\thanks{Associate Professor, Director, Land, Air and Space Robotics (LASR) Laboratory, Aerospace Engineering},
\\
\textit{Texas A\&M University, College Station, TX, 77843-3141}}

\maketitle{} 		

\section {Introduction}
The explosive growth of space technologies that are furthering the information age continue to elicit transformative advances that can chiefly be attributed largely to the inexpensive and reusable commercial launch capabilities. They form a basis for the space technology of the future, which include on-orbit servicing, assembling and manufacturing. Next generation space robotics technologies that focus on particular mechanisms and machines that are suitable for operations in micro and nano gravity environments form an important aspect of research. The confluence of advanced manufacturing technologies, novel materials, electronics and advanced computing architectures forms a fertile precursor for this research. One of the most versatile set of space mechanisms that can be a potential enabler are free-flying systems. Advances in CubeSat technology have continued to further the technology maturation of the free-flyer concept. Ground robotic platforms that focus on CubeSat technology enable accessible and low-cost spacecraft motion emulation solutions that contrast industrial and defense standards \cite{wilde2019historical}. This leads to regularized hardware in the loop testing of free-flying space robotic mechanisms\cite{Heidt2000CubeSatAN} that build confidence in novel small space system mission concepts.


The Land, Air, and Space Robotics (LASR) Laboratory at Texas A\&M University has made significant contributions in the areas of dynamics, control and simulation of a Mobile Robotic System for 6-DOF Motion Emulation\cite{LASR_HOMER}. The laboratory is home to a variety of robotic platforms that emulate satellite motions \cite{HOMER_odo, HOMER_proxops}. One key robotic testbed is the Holonomic Omnidirectional Motion Emulation Robot (HOMER) that not only replaces air bearing tables for robotic proximity operation testing, but offers realistic 6-DOF motions to test sensors and actuators. The LASR Laboratory is now investigating the use of free-flying spacecraft modules in several different on-orbit, servicing and manufacturing (OSAM) activities as part of the Space University Research Initiative (SURI) funding opportunity awarded by the Air Force Office of Scientific Research to the "Breaking the 'Launch Once, Use Once' Paradigm" Team, which consists of Carnegie Mellon University (CMU), University of New Mexico (UNM), Texas A\&M University (TAMU), and the Northrop Grumman Corporation (NGC). The work done by TAMU in the LASR Laboratory for this SURI project involves the system development and testing of the aforementioned thrust-capable modules.


This paper presents a detailed system design and component selection for the Transforming Proximity Operations and Docking Service (TPODS) module, designed to gain custody of uncontrolled resident space objects (RSOs) via rendezvous and proximity operation (RPO). In addition to serving as a free-flying robotic manipulator to work with cooperative and uncooperative RSOs, the TPODS modules are engineered to have the ability to cooperate with one another to build scaffolding for more complex satellite servicing activities. The structural design of the prototype module is inspired by Tensegrity principles \cite{skelton2001introduction}, minimizing the structural mass of the module's frame. The prototype TPODS module is fabricated using lightweight polycarbonate with an aluminium or carbon fiber frame. The inner shell that houses various electronic and pneumatic components is 3-D printed using ABS material. Four OpenMV H7 R1 cameras are used for the computer vision/pose estimation of resident space objects (RSOs), including other TPODS modules. Based on anticipated system dynamics and mass properties, a comprehensive propulsion system analysis is performed for subsequent hardware selection. Compressed air supplied by an external source is used for the initial testing and can be replaced by module-mounted nitrogen pressure vessels for full on-board propulsion later. A Teensy 4.1 single-board computer is used as a central command unit that receives data from the four OpenMV cameras, and commands its thrusters based on the control logic. Extensive simulations are performed spanning multiple control strategies to enable closed-loop tracking of a commanded trajectories. A metric to asses the close loop tracking of commanded input and duration of thruster firing is calculated in each case, and the optimal control strategy is chosen. For the final validation of the integrated system, a near-frictionless testing environment is created with spherical ball transfers atop a plexiglass workspace. The performance of the pose estimation algorithm and control system with fuel and time constraints is validated via tracking a moving target.

In its contribution to the SURI project, the LASR Laboratory's technological development procedure works to produce systems that are only a few steps away from real hardware implementation and spaceflight. Because of this notion, the prototype TPODS module sizing was meant to fit a 1U CubeSat silhouette. This follows a goal of fitting all/most necessary subsystems in the smallest possible payload, with the idea in mind that scaling up after this is achieved is more or less trivial. TPODS modules are intended to be launched from a host chaser spacecraft and attach onto a target body via electromagnetics, GeckSkin\cite{patek2014biomimetics}, or other non-damaging adhesives. The modules will impart an impact, stick, and thrust (in moment couples) combination to stabilize the target's spin rates \cite{Down:2023}. Once on the target body, modules will reconfigure themselves into scaffolding structures to facilitate servicing activities by the host chaser. In accomplishing these tasks, both module to module, and chaser to module pose estimation solutions are necessary, in addition to the target's pose estimation by the chaser.

\section{System Dynamics}
The overall behavior of the satellite module is significantly dependent on the location and arrangement of the applied reaction force via thrusters. In general, to enable $n$ Degrees Of Freedom (DOF) holonomic control of a rigid body, $n+1$ unidirectional actuators are required\cite{10.2307/2372648}. However, for the practical applications, configurations with multiple redundant thrusters have been employed for greater flexibility and robustness against complete or partial failure of single or multiple thrusteres\cite{Astrobee}. The overall goal of this design experiment is to demonstrate and validate the concept of operation with the bare minimum system complexity. Consequently, it was decided to incorporate four independent unidirectional thrusters (for three DOF holonomic motion) in the prototype module. The following few subsections contain an exhaustive analysis of the impact of various thruster configurations on the overall behavior of the satellite module.  

\subsection{Configuration of On-board Thrusters}
\begin{figure}[t!]
     \centering
     \begin{subfigure}[b]{0.35\textwidth}
         \centering
         \includegraphics[width=\textwidth]{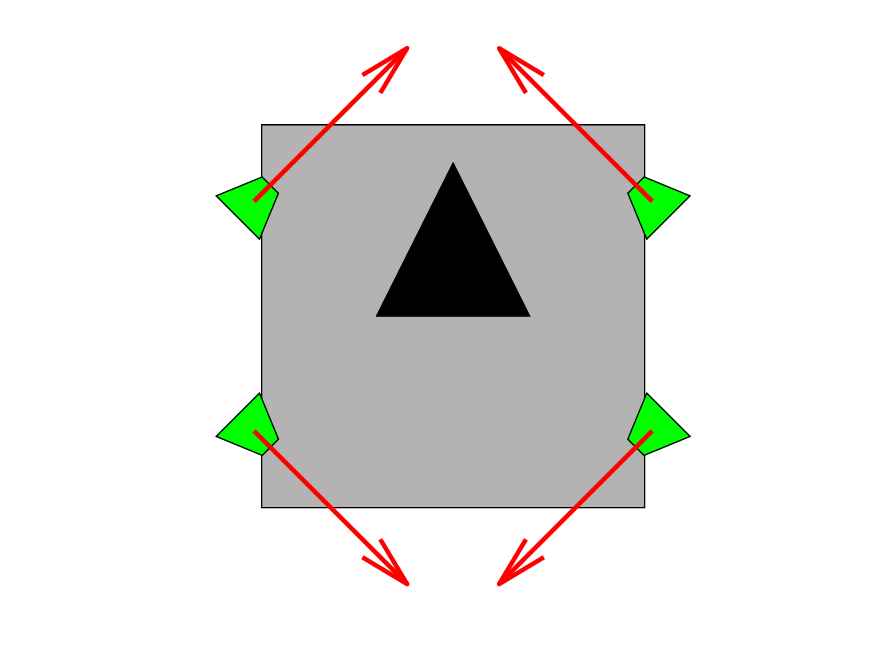}
         \caption{X Configuration}
         \label{x}
     \end{subfigure}
     \hfill
     \begin{subfigure}[b]{0.28\textwidth}
         \centering
         \includegraphics[width=\textwidth]{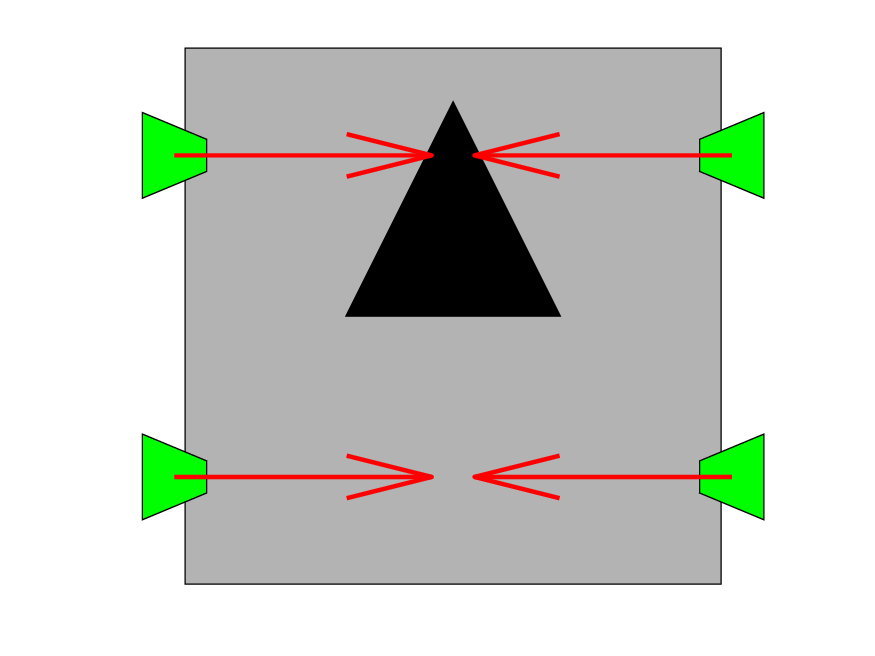}
         \caption{H Configuration}
         \label{h}
     \end{subfigure}
     \hfill
     \begin{subfigure}[b]{0.32\textwidth}
         \centering
         \includegraphics[width=\textwidth]{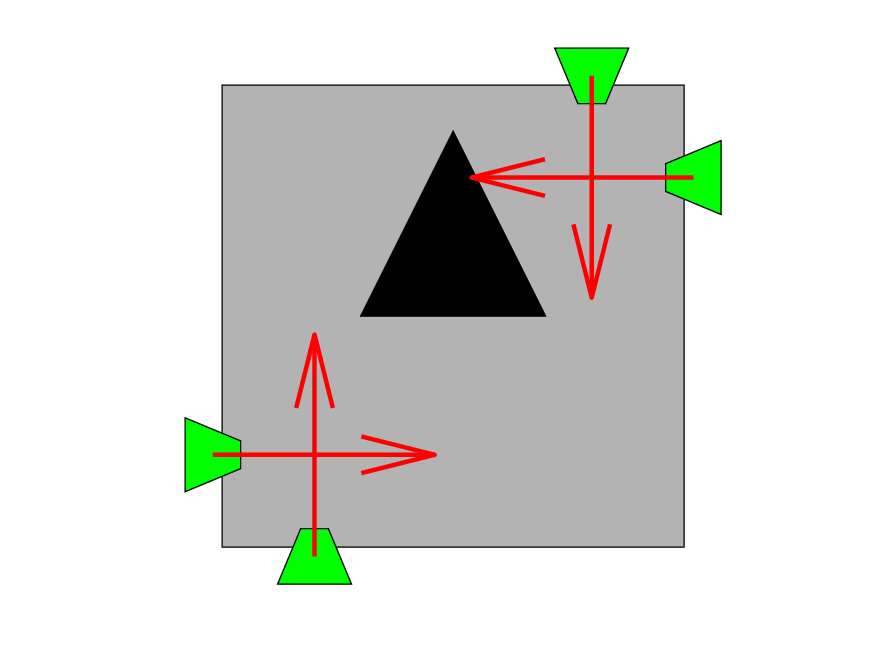}
         \caption{Offset H Configuration}
         \label{hoffset}
     \end{subfigure}
        \caption{Various thruster configurations for the satellite module}
        \label{fig:thruster_config}
\end{figure}

A fixed number of thrusters on the module can be arranged in distinct ways for specific mission objectives. For this design experiment, several possible configurations of the four thrusters have been studied, and the ones shown in Figure \ref{fig:thruster_config} are analyzed in more depth. The $H$ configuration shown in Figure \ref{h} appears to be a viable choice for the unidirectional translation and rotational motions. However, the translation motion in the axis perpendicular to the thrust line can not be achieved without the rotation of the module. This is a significant limitation for an application involving proximity operations and rendezvous of the module with another module or body. The \textit{offset H} configuration enables pure sideways motion by a combination of two perpendicular thrusters. However, it is not possible to get the translation motion in the direction of the thrust line of the thrusters. The $X$ configuration, as shown in Figure \ref{x}, enables independent translation motion in the direction of the principal perpendicular axis as well as in the direction of the thrust line of the thrusters. It also provides excellent control authority of the orientation angle due to the large moment arm. This configuration provides a reasonable holonomic motion due to the decoupling of translation and rotational motion; and has been extensively studied and employed for attitude stabilization during the interplanetary EDL phase\cite{MSL}. Consequently, the $X$ thruster configuration is selected for the TPODS modules.   

\subsection{Equations of Motion}
For the analysis of the motion of the module under the influence of various thrust inputs, a Cartesian coordinate system centered at the geometric center of the module and axes parallel to the subsequent side has been considered. The X axis is towards the the page's right, and Y axis is towards the top. With these reference axes and the physical properties of the satellite module, as mentioned in Table \ref{table:1}, the motion governing equations can be analyzed considering the state, input, and output variables presented in Table \ref{table:2}.

\begin{table}[h!]
\centering
\parbox{0.3\textwidth}{
\begin{tabular}{|l|l|} 
 \hline
 Property & Value \\ 
 \hline
 Mass & 2.268 kg  \\ 
 Length L & 10 cm \\
 Moment Arm d & 5 cm \\
 Moment of Inertia & \multirow{2}{*}{$\frac{mL^2}{6}$} \\
 along z axis $I_{zz}$ &  \\
 \hline
 \end{tabular}
 \caption{Physical Properties of TPODS.}
 \label{table:1}
 }
\hfill  
\parbox{0.65\textwidth}{
\begin{tabular}{|l|l|l|} 
 \hline
 States & Input & Output\\
 \hline
  X position \textbf{x} & Thruster $\bm{T_1}$ & X position \textbf{x}\\ 
  Y position \textbf{y} & Thruster $\bm{T_2}$ & Y position \textbf{y}\\
  Orientation angle $\bm{\Psi}$ & Thruster $\bm{T_3}$ & Orientation angle $\bm{\Psi}$\\
  X speed $\bm{u}$ & Thruster $\bm{T_4}$ & \\
  Y velocity $\bm{v}$ & &\\
  Angular rate $\bm{r}$ & & \\
 \hline
 \end{tabular}
\caption{States, Input and Output of the TPODS plant.}
\label{table:2}}
\end{table}

Considering the planner motion of the module and the arrangement of the thrusters, it follows standard 3-DOF equations of motion in the body attached frame are\cite{nelson1998flight} 
\begin{align} 
\dot{x} &= u\label{eq:EOM1}\\
\dot{y} &= v\\
\dot{\Psi} &= r\\
\dot{u} &= \frac{\Sigma{F_x}}{m}+rv = \frac{\left(T_{2}+T_{3}-T_{1}-T_{4}\right)}{m\sqrt{2}} + rv\\
\dot{v} &= \frac{\Sigma{F_y}}{m}-ru = \frac{\left(T_{1}+T_{2}-T_{3}-T_{4}\right)}{m\sqrt{2}} - ru\\
\dot{r} &= \frac{\Sigma{M_y}}{I_{zz}} = \frac{d\left(T_{1}+T_{3}-T_{2}-T_{4}\right)}{I_{zz}}\label{eq:EOM2}
\end{align}

\begin{figure}[b]
    \centering
    \includegraphics[width=0.54\textwidth]{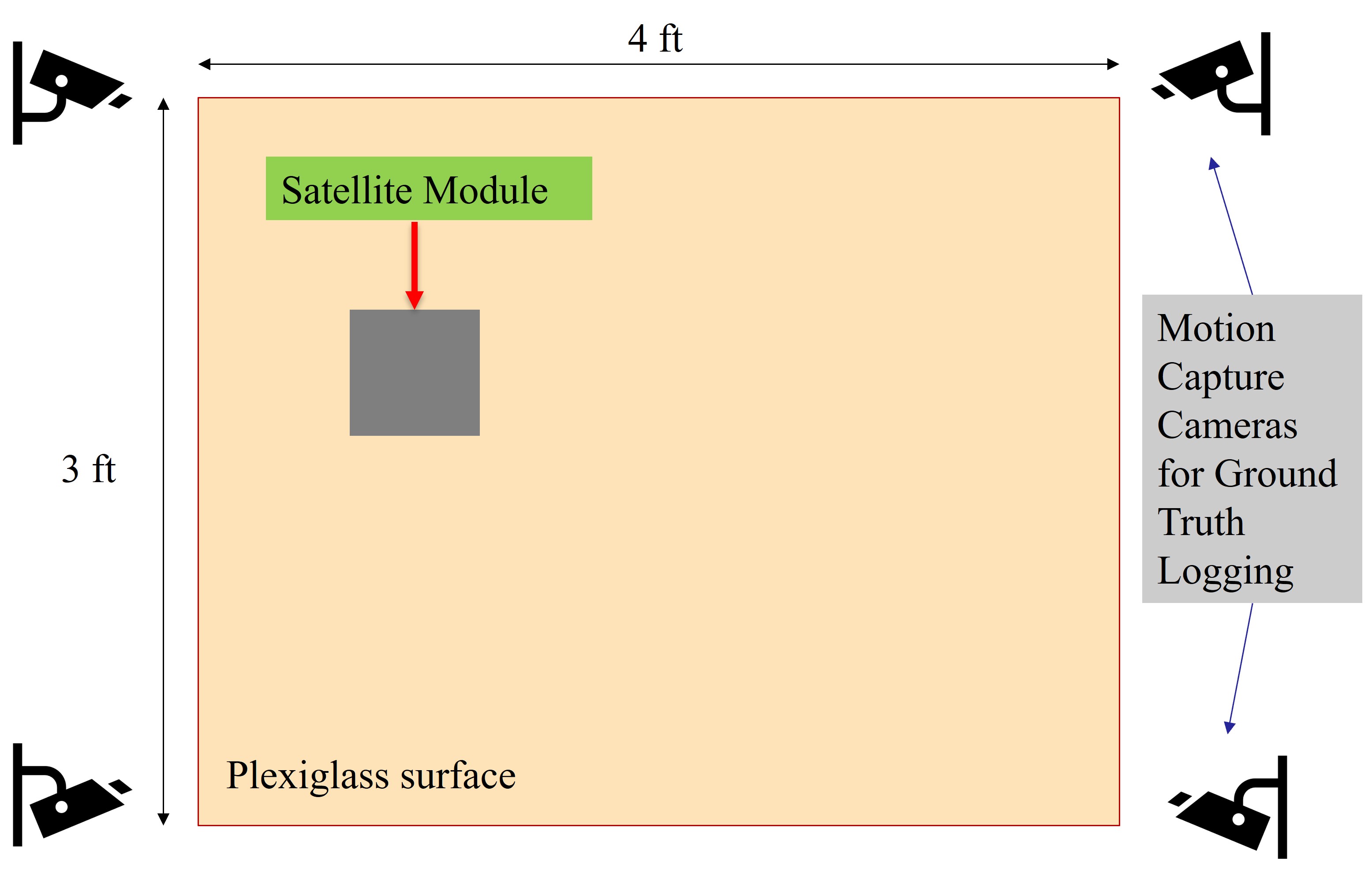}
    \includegraphics[width=0.45\textwidth]{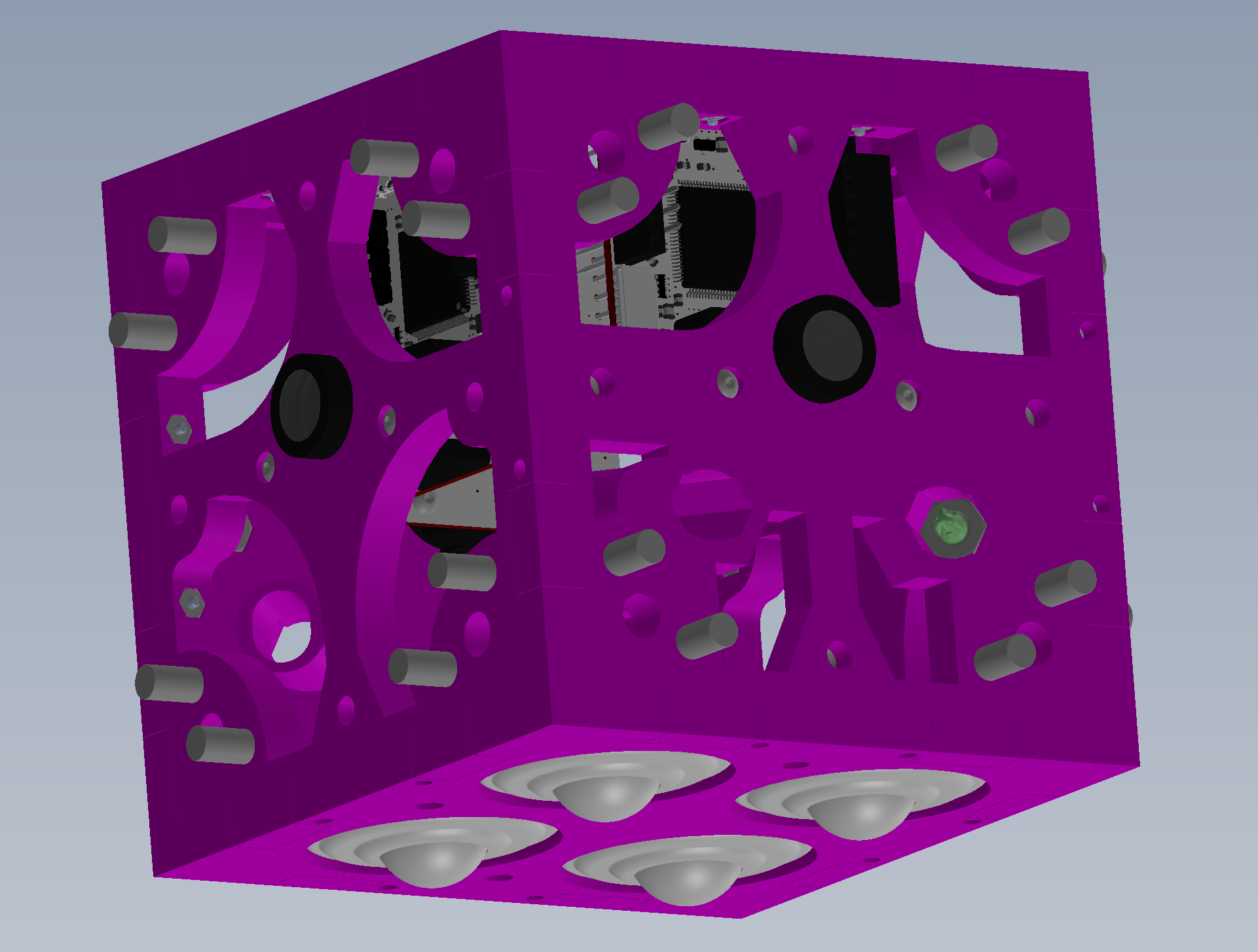}
    \caption{Tabletop experiment setup and TPODS module}
    \label{fig:3}
\end{figure}

Since Equations \ref{eq:EOM1}-\ref{eq:EOM2} describes a non-linear system, a linearization of the system described by Equations \ref{eq:EOM1}-\ref{eq:EOM2} has to be performed around an operating point to conduct the primary open loop analysis of the system. The linearization process involves defining a new set of variables for the states, inputs, and outputs; centered on the operating point. The Jacobians for linearized state-space equations in terms of variance can be derived next with the new set of variables. With the process mentioned above, the Jacobians are computed and evaluated at the operating conditions $u_0,v_0,r_0=0$. As presented in Figure \ref{fig:3}, the overall goal of this design exercise is to devise, build and validate autonomous satellite modules with subsequent vision-based pose estimation and control strategies using cold gas thrusters. Considering the deliverable, the following analysis has been carried out to validate the effectiveness of the thruster configuration and gauge the operating envelope of the satellite module.

\section{Design of Closed-loop Controller}
\subsection{Controllability and Analysis}
From the linearized system dynamics matrix, it can be shown that irrespective of the operating conditions of the position and orientation, all eigenvalues of the system dynamics matrix are zero. Since all modes of the system are unstable, there are no damping ratio and time constants applicable. The controllability matrix is full rank for all operating conditions of position and orientation. Without loss of generality, assume that the T1 thruster has failed. With the failed thruster, the controllability matrix still has the full rank; hence the system is \textit{reachable}. This result is expected as any combination of two opposite thrusters, and one adjacent thruster can still control the planner motion of the satellite. At any position, the pair of opposite thrsters can be used to achieve pure rotation and the remaining adjacent thruster can provide restoring moment. However, loss of one thruster results into loss of independent control of position and orientation. Further, Without loss of generality, assume that T1 and T2 thrusters have failed. In this case, the controllability matrix has the rank 4, which is less than the dimension of the system; hence the overall system with two failed thrusters is \textit{NOT reachable}.

\subsection{Desired Closed-Loop Characteristics and Sampling Time}
Table \ref{table:3} presents desired closed-loop performance parameters. They are selected such that the overall system exhibits a highly damped response. Since the total amount of onboard propellant is limited, it is desired to have a slow but fuel-efficient response without much oscillation or overshoot. There is always an option to make the overall system faster by setting more aggressive gains. However, such a response tends to consume more propellant than the highly damped one mentioned earlier.

\begin{table}[h]
\centering
\begin{tabular}{|l | l|} 
 \hline
 Property & Value \\ 
 \hline
 Rise Time & 4 sec  \\ 
 Settling Time & 8 sec \\
 Maximum Peak Overshoot & 10 \\
 Damping Ratio & 0.8 \\
 \hline
 \end{tabular}
 \quad
\begin{tabular}{|l | l|} 
 \hline
 Property & Value \\ 
 \hline
 Rise Time & 8 sec  \\ 
 Settling Time & 10 sec \\
 Maximum Peak Overshoot & 5 \\
 Damping Ratio & 0.9 \\
 \hline
 \end{tabular}
\caption{Desired time domain specifications for Closed-loop Orientation and Position Controller}
\label{table:3}
\end{table}

Since all system modes are \textit{unstable}, no damping ratios and time constants are applicable to find the Nyquist frequency and sampling time. Since the position and orientation of the satellite module are estimated by the computer vision algorithm, sample period of \textit{0.1s} is selected for the \textit{on-orbit operations}. However, for the tabletop experiment, high-speed position and orientation data are available from the motion capture system. Hence, a sample period of \textit{0.02s} is selected for the \textit{simulation and tabletop experiment}\cite{6U}. 

\subsection{Sampled Data Regulator with Integral and Non-Zero Set-point Feed-forward Terms}
\subsubsection{Discrete-Time State Space Model}
: MATLAB{\textregistered} command \textit{c2d} with sample time 0.1s is used to discretize the linearized continuous time transfer function considering the first order lag dynamics with the time constant of 0.1 seconds for thrusters. The actuator dynamics, combined with the original state equations, form a state space model with thrust commands as the inputs and actual thrust as additional states. 

\subsubsection{Closed-loop Controller design and verification :} \label{sssec:num1}
\begin{algorithm}
\caption{General process to calculate Feedback Gain for SDR}\label{alg:SDR}
\begin{algorithmic}[1]
\State Discretize the state space model : $\dot{x} = Ax + Bu \to x((k+1)T)=G(T)x(kT)+H(T)u(kT)$
 Where, $G(T) = e^{AT}$, $H(T) = \int_{0}^{T} e^{AT} B \,d\tau$, $u(t) = u(kT), kT\leq t < (k+1)T$ 
\State Identify the performance index to be minimized : J = $\frac{1}{2}\int_{0}^{NT}[x'(t)Qx(t)+u'(t)Ru(t)]\,dt$
\State Rewrite : $\dot{x}(t) = \zeta(t-kT)x(kT) + \eta(t-kT)u(kT)$
 Where, $\zeta(t-kT) = e^{A(t-kT)}$ and $\eta(t-kT)=\int_{kT}^{t} \zeta(t-kT) B \,d\tau$ 
\State $Q_1 \gets \int_{kT}^{(k+1)T} Q\zeta^2(t-kT)\,dt$
\State $M_1 \gets \int_{kT}^{(k+1)T} Q\zeta(t-kT)\eta(t-kT)\,dt$
\State $R_1 \gets \int_{kT}^{(k+1)T} [Q\eta^2(t-kT)+R]\,dt$
\State $\hat{Q} = Q_1 - M_1R_1^{-1}M_1'$
\State $v(k) = R_1^{-1}M_1'x(k) + u(k)$
\State Substitute Step 4 to 8 into Step 2 : J = $\frac{1}{2}\sum_{k=0}^{N-1}[x'(k)\hat{Q}x(k)+v'(k)R_1v(k)]$  
\State Substitute Step 7 into Step 1 : $x(k+1)=\hat{G}x(k)+Hv(k)$ Where, $\hat{G}=G-HR_1^{-1}M'_1$
\State Quadratic performance index of Step 9 along with the system dynamics of Step 10, forms a standard DLQR problem.  
\end{algorithmic}
\end{algorithm}

The discrete state space model is further utilized to formulate a standard Sampled Data Regulator (SDR) problem\cite{ATHANS}. Since the discrete Linear Quadratic Regulator (DLQR) penalizes any deviations in states and control variables at multiple instances of the sample time only, the overall gains resulted from DLQR do not account for the continuous nature of system dynamics and the cost function. Contrary to this, the SDR formulation acts upon the continuous cost function, resulting in more realistic gain values required to achieve the desired closed-loop response. The high-level process to compute the feedback gain is summarized in the Algorithm \ref{alg:SDR}\cite{ogata-1987a }.

\begin{figure}[h!]
\centering
\includegraphics[width=0.49\textwidth]{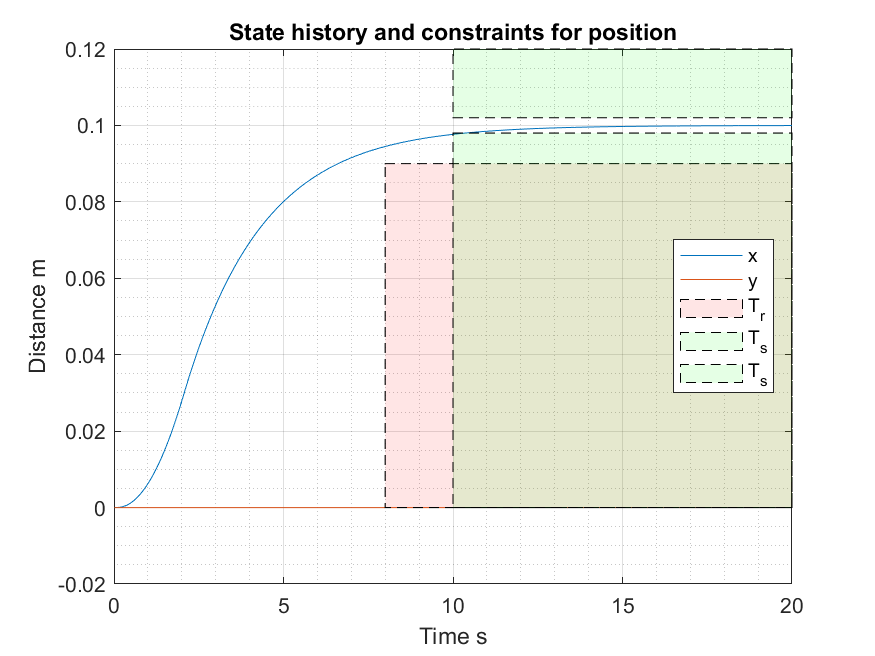}
\includegraphics[width=0.49\textwidth]{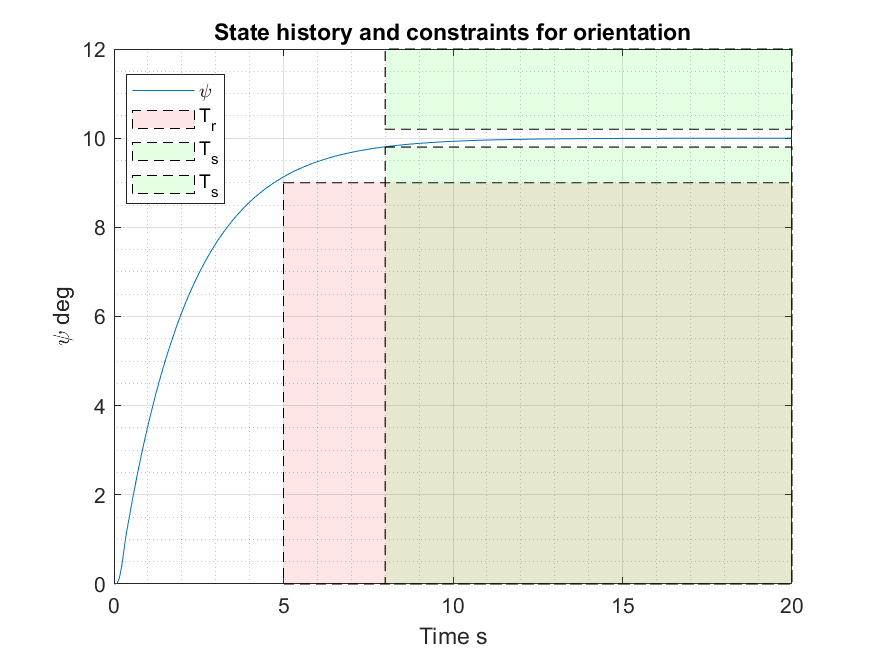}\\
\caption{Simulation Histories of $x$ and $\psi$ with rise time $T_r$ and settling time $T_s$ requirements for TPODS planar motion with SDR controller. ($T_r$ = 8 sec and $T_s$ = 10 sec for position, $T_r$ = 5 sec and $T_s$ = 8 sec for the orientation)}
\label{Q741_const}
\end{figure}

\begin{figure}[p]
\centering
\includegraphics[width=0.45\textwidth]{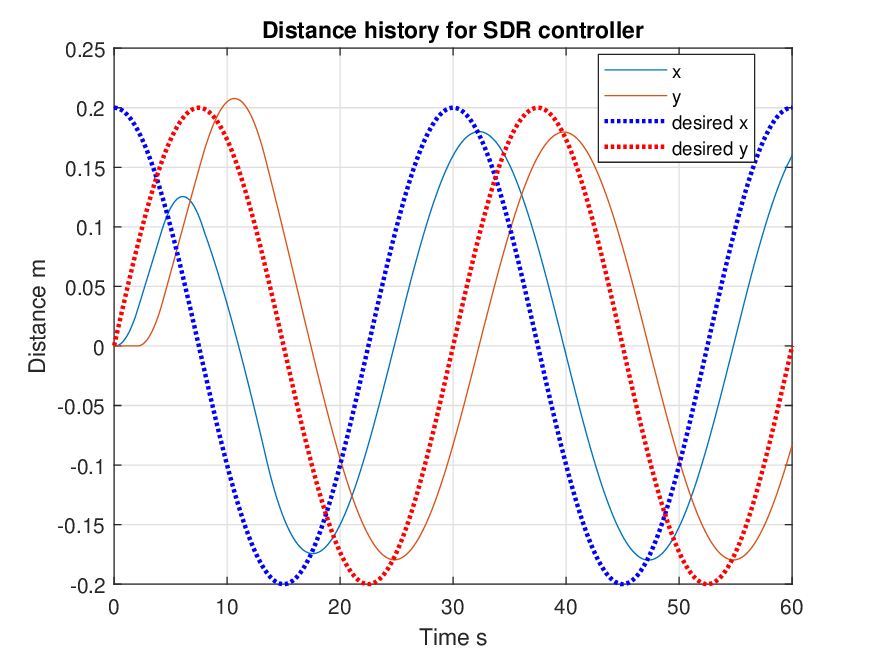}
\includegraphics[width=0.45\textwidth]{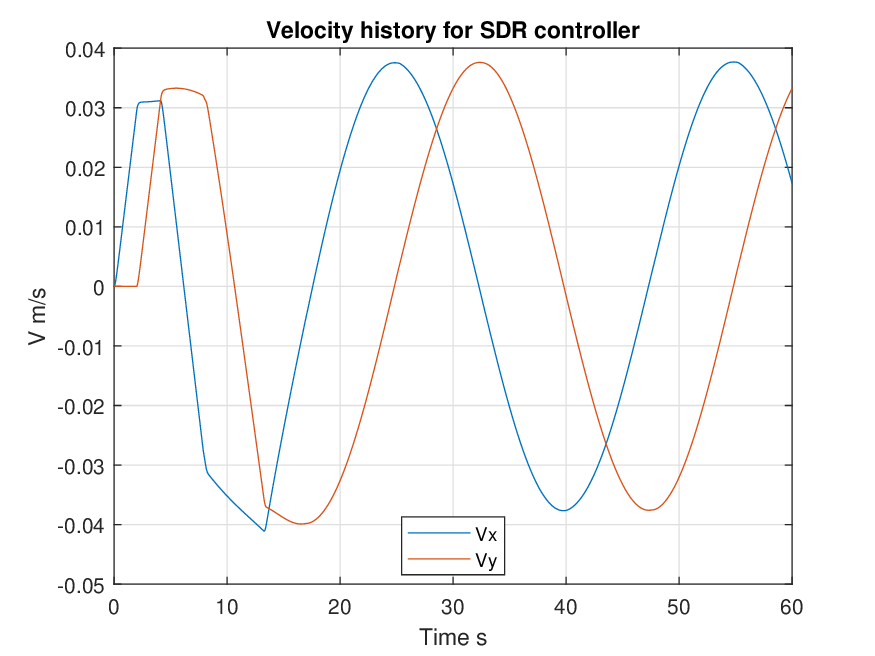}\\
\includegraphics[width=0.45\textwidth]{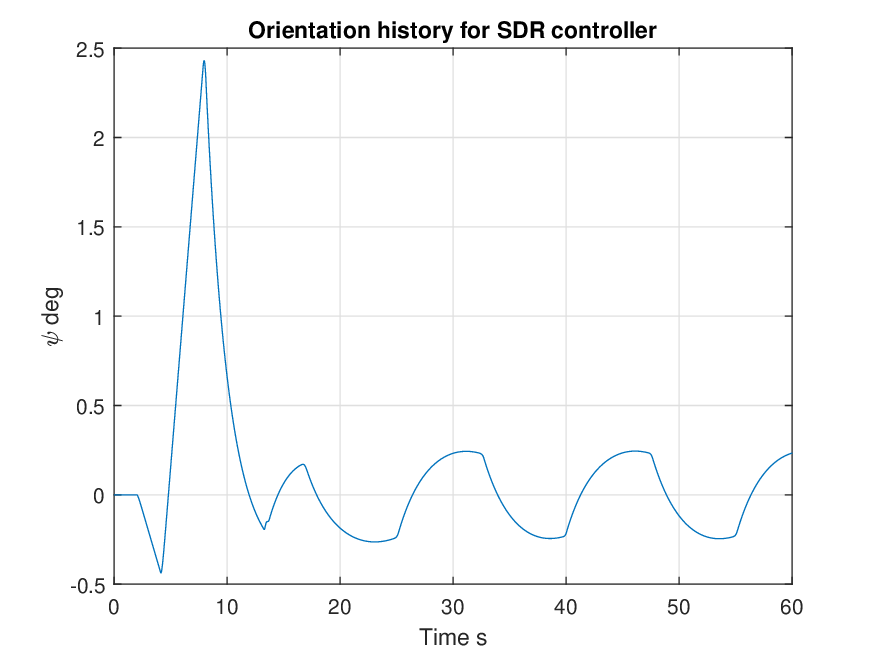}
\includegraphics[width=0.45\textwidth]{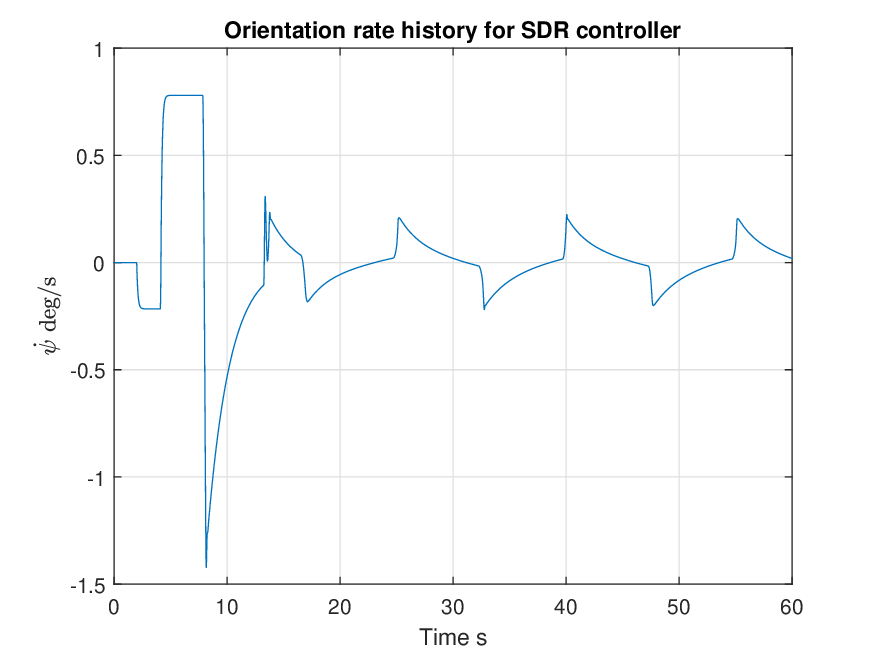}\\
\includegraphics[width=0.85\textwidth]{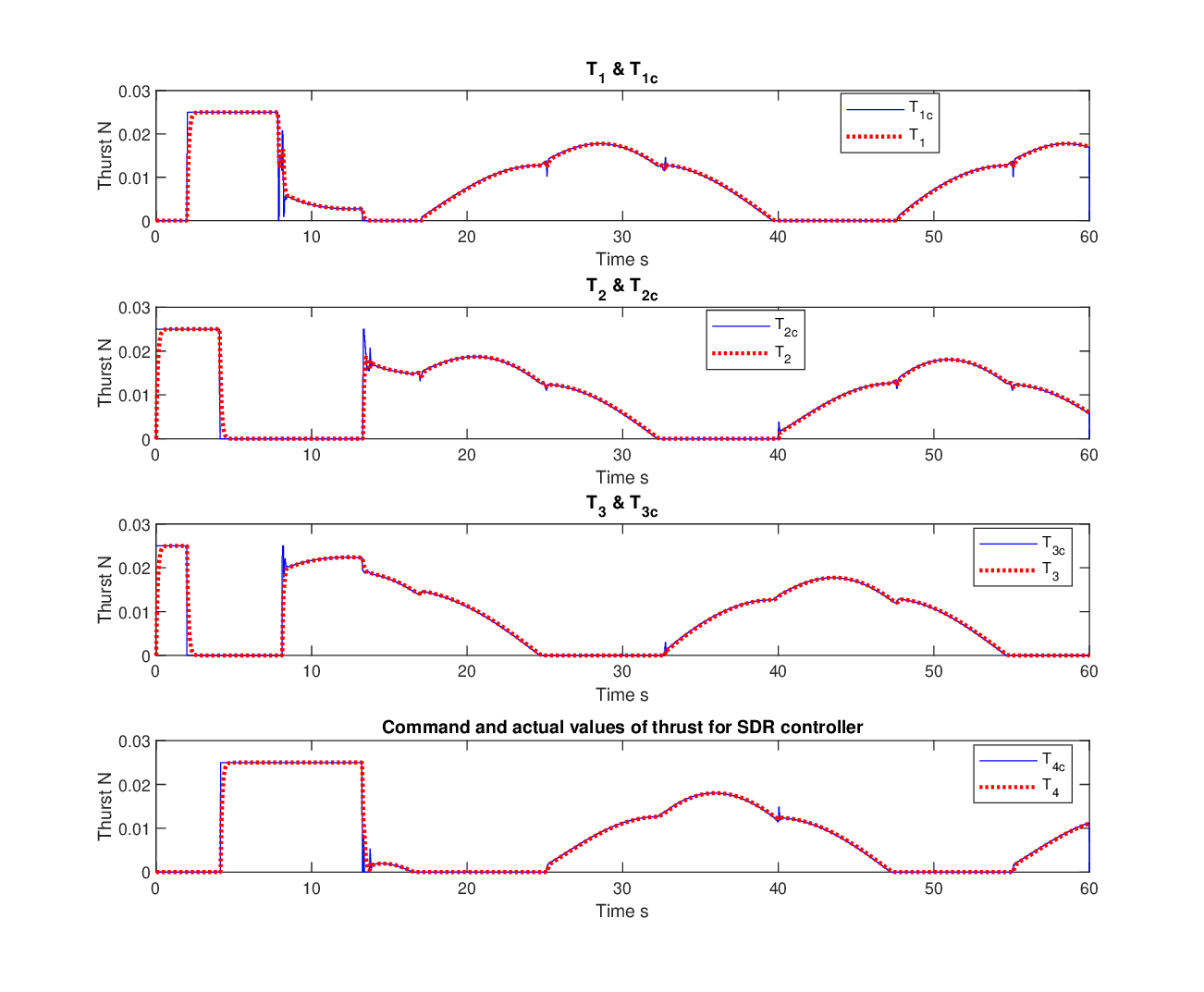}
\caption{Simulation histories of states, commanded control inputs $T_c$, and actual control inputs T for sinusoidal tracking signal of magnitude $0.2$ m and period of $30$ sec  in $x$ and $y$ with SDR controller}
\label{Q76}
\end{figure}

Once the SDR formulation is converted into the standard DLQR form, algebraic matrix Riccati equation can be utilized to compute the steady-state feedback gains for the full-state feedback controller. The weighing matrices Q and R are computed using Bryson's rule, and the system performance for computed state feedback gains is evaluated using a discrete plant model with the integration step of 10ms. It can be concluded from the Figure \ref{Q741_const} that a well-tuned controller meets all the performance requirements for the closed-loop system as envisioned in Table \ref{table:3}. The states and control history of a sinusoidal input of 0.2 m magnitude and period of 30 seconds is shown in Figure \ref{Q76}. It is evident from the state history that while the controller can follow the general nature of the sinusoidal input, there is a constant error between the current state and the desired state. This is expected as the feedback regulator does not have any integral term as well as any prior information about the desired state. 

\subsubsection{Integral and Non-Zero Set-point Feed-forward Terms :}
\begin{figure}[h!]
\centering
\includegraphics[width=\textwidth]{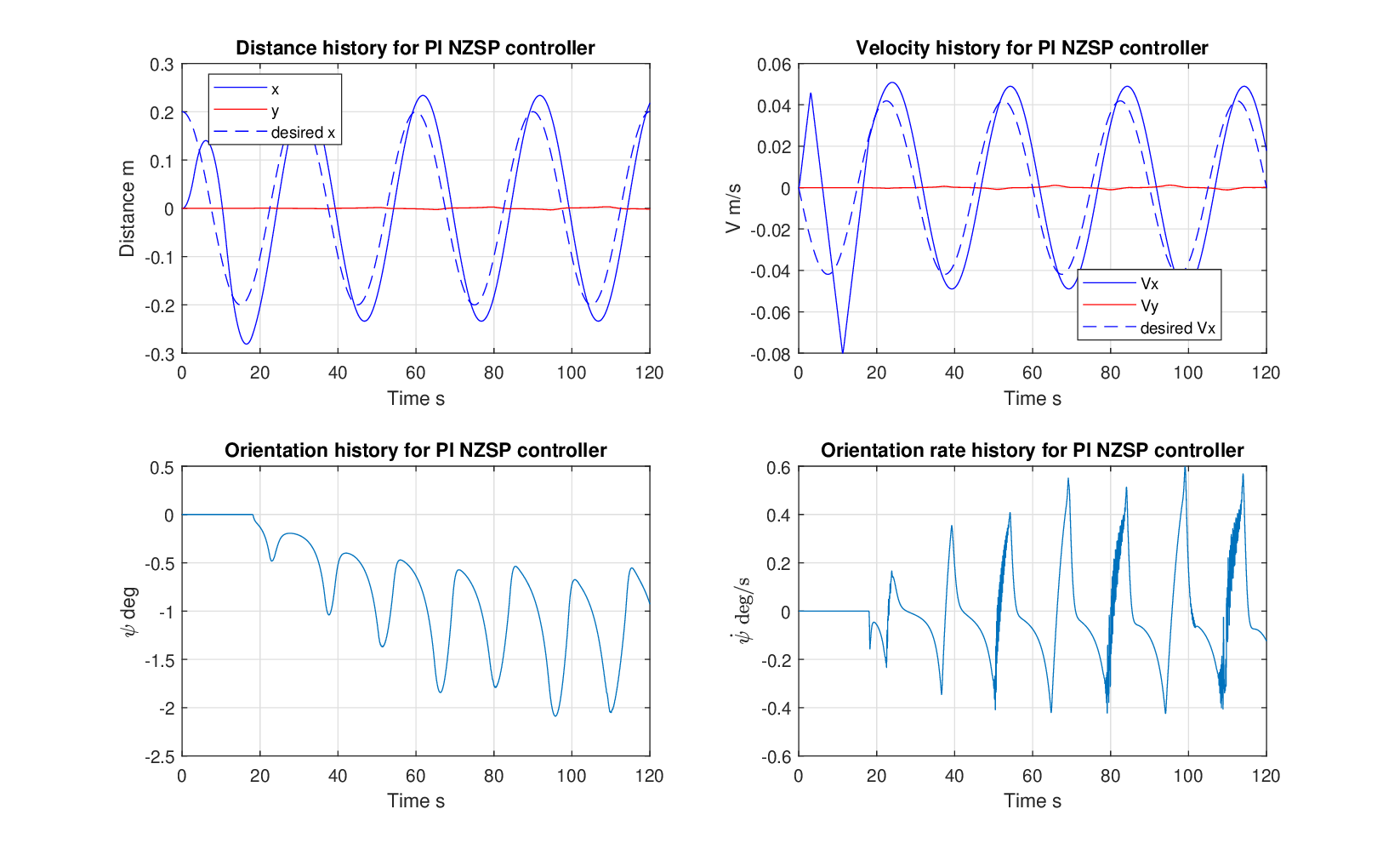}
\caption{Simulation Histories of states for sinusoidal input of magnitude $0.2$ m and period of $30$ sec  in $x$ and $y$ with PI-NZSP SDR controller}
\label{PINZSPSDR}
\end{figure}

The SDR formulation does not contain any mechanism to eliminate steady-state errors. Such behavior is fatal in the presence of external process disturbances. Adding an integral term associated with states of interest provides a robust solution for disturbance rejection. This can be achieved by augmenting the state vector with the integrated states. The exact process of Algorithm \ref{alg:SDR} can be then followed with an augmented state space model to calculate feedback gains. Further, the output always lags behind the desired dynamics input due to the inherent nature of the delay associated with the feedback loop. In addition, for most practical systems, states are commanded to follow a specific trajectory of non-zero set points instead of driving states to zero. For such scenarios, a feed-forward path, which pre-calculates the required control inputs for the given desired state trajectory, enables fast and reliable tracking of non-zero set points. The pre-calculation of the control inputs relies on the dynamics of the system. The contribution of the feed-forward term in the overall control input depends on the desired set-point and elements of the quad-partition matrix, which essentially captures the dynamic behaviour of the system. As summarized in Algorithm \ref{alg:PI-NASP} both formulations can be combined to achieve reliable tracing of non-zero set-points and robust disturbance rejection. The states and control history of a sinusoidal input of 0.2 m magnitude and period of 30 seconds is shown in Figure \ref{PINZSPSDR}. It is evident from the state history that the controller can follow the general nature of the sinusoidal input, matching amplitude and eventually reaching to the commanded value in a particular cycle. 

\begin{algorithm}
\caption{Sampled Data Regulator with Integral and Non-Zero Set-point Feed-forward Terms}\label{alg:PI-NASP}
\begin{algorithmic}[1]
\State Augment the integral term in original plant model and compute higher dimensional feedback gain for the original system using  higher dimensional $Q_1$ matrix. 
\State Discrtize the original unaugmented state space model : $\dot{x} = Ax + Bu \to x((k+1)T_c)=G_c(T_c)x(kT_c)+H_c(T_c)u(kT_c)$
 Where, $G_c(T_c) = e^{AT_c}$, $H_c(T_c) = \int_{0}^{T_c} e^{AT_c} B \,d\tau$, $u(t) = u(kT_c), kT_c\leq t < (k+1)T_c$ 
 \Ensure The sampling time used to discritize system $T_c$ depicts the rate at which the feedback controller computes the control inputs. This is usually different than the sampling time T = 10ms which is used to validate the performance of the closed-loop discrete system.  
 \State Compute the sub-matrices $\pi_{12}$ and $\pi_{22}$ of the quad partition matrix using \newline
 $
\left[ \begin{array}{c:c}
G_c-I & H_c \\ \hdashline
C & D
\end{array}\right]
= 
\left[ \begin{array}{c:c}
\pi_{11} &  \pi_{11} \\ \hdashline
\pi_{21} & \pi_{22}
\end{array}\right]
$ Where the goal is to drive $y=Cx+Du$ to $y_m$
\State Validate the performance of the closed-loop system using non zero $y_m$ and control law \newline $u_k = (\pi_{22}+K_1\pi_{12})y_m-Kx_k$. Where $K = [K_1 K_2]$, $K_1$ represents the feedback gain for unsegmented system and $K_2$ is the gain associated with the integral terms. 
\end{algorithmic}
\end{algorithm}

\section{Fuel Optimal Trajectories}
While a well-tuned SDR can drive the current state to the desired state through a near extremal trajectory \cite{dorato}, it is often not trivial to end up with an optimal set of gains. Instead, the SDR formulation guarantees that for the given set of $Q$ and $R$ matrices, the computed gain results in a trajectory that minimizes the cost function defined by deviations in states and control inputs. Hence it is often necessary to perform a comprehensive analysis driven by a calculus of variations to gain insights into extremal trajectories for a given dynamical system. Such exercise involves the process summarized in Algorithm \ref{alg:optimal} followed by the iterative numerical solution of a two-point boundary value problem, provided applicable boundary conditions, to arrive at the optimal time histories of necessary control inputs and resulting states. 

\begin{algorithm}
\caption{General process to formulate an optimal control problem}\label{alg:optimal}
\begin{algorithmic}[1]
\State For a system with states $x$ and controls $u$, select cost function to be optimized $J(x,u,t)$
\State Using the governing equations of states and the cost function, formulate Hamiltonian $H(x,u,t,\lambda)$, where $\lambda$ are co-state variables.
\State Apply necessary conditions for optimality $\frac{\partial H}{\partial u}=0$ and compute extermal input $u^*(x,\lambda,t)$
\State Rewrite the Hamiltonian by replacing $u$ with the extremal  input $u^*$
\State Applu Euler-langrage equations to compute state equations $\dot{x} = \frac{\partial H}{\partial \lambda}$ and costate equation $\dot{\lambda}=-\frac{\partial H}{\partial x}$
\State Incorporate any state and/or output inequality constraints.
\State Formulate and solve two point boundary value problem enforcing applicable boundary conditions.
\end{algorithmic}
\end{algorithm}

For the motion of the satellite module in a plane with 3-DOF, trajectories that result in the least energy consumption are the primary interest of the analysis presented in this section. Consequently, the quadratic cost function of Equation \ref{cost_fcn}, which only depends on the inputs, is chosen. Since, the time for the motion is assumed to be given and constant, the cost function does not depend on the final time of the motion, $t_f$.       
\begin{equation}
J =  \int_{0}^{t_f} \frac{1}{2} (T_1^2+T_2^2+T_3^2+T_4^2)\,dt 
\label{cost_fcn}
\end{equation}

The Hamiltonian for the cost function of Equation \ref{cost_fcn} and dynamical governing Equations \ref{eq:EOM1}-\ref{eq:EOM2} can be written as\cite{kirk2012optimal}
\begin{align}\label{H}
H &=  \frac{1}{2} (T_1^2+T_2^2+T_3^2+T_4^2)+\lambda_1u+\lambda_2v+\lambda_3r+\lambda_4\left(\frac{\left(T_{2}+T_{3}-T_{1}-T_{4}\right)}{m\sqrt{2}} + rv\right) \nonumber \\ 
&+\lambda_5\left(\frac{\left(T_{1}+T_{2}-T_{3}-T_{4}\right)}{m\sqrt{2}} - ru\right)+\lambda_6\left(\frac{d\left(T_{1}+T_{3}-T_{2}-T_{4}\right)}{I_{zz}}\right)
\end{align}
\subsection{Necessary and Sufficient conditions for an extremal  trajectory}
With the Hamiltonian defined for the system, the necessary conditions for an extremal trajectory, $H_u=0$, can be enforced to determine the optimal control inputs\cite{kirk2012optimal}. For the Hamiltonian given by Equation \ref{H}, application of necessary conditions results in the following optimal control inputs
\begin{align}\label{optimal_u}
T_1^* &= \frac{1}{\sqrt{2}m}(\lambda_4-\lambda_5)-\frac{d}{I_{zz}}\lambda_6 \qquad
T_2^* = \frac{1}{\sqrt{2}m}(-\lambda_4-\lambda_5)+\frac{d}{I_{zz}}\lambda_6 \nonumber\\
T_3^* &= \frac{1}{\sqrt{2}m}(\lambda_5-\lambda_4)-\frac{d}{I_{zz}}\lambda_6 \qquad
T_4^* = \frac{1}{\sqrt{2}m}(\lambda_4+\lambda_5)-\frac{d}{I_{zz}}\lambda_6 
\end{align}

Additionally, it can be proven that the extremal trajectory given by the optimal inputs of Equations \ref{optimal_u} minimizes the cost function as the Legendre-Clebsch condition, $H_{uu} > 0$, is always true for the given formulation. With the optimal inputs given by Equations \ref{optimal_u} the Hamiltonian of Equation \ref{H} can be rewritten by eliminating the control inputs as,

\begin{equation}\label{H_revised}
H = \lambda_1u+\lambda_2v+\lambda_3r+\frac{\lambda_4^2}{m^2}+\lambda_4rv-\frac{\lambda_5^2}{m^2}-\lambda_5ru-\frac{2d^2}{I_{zz}^2}\lambda_6^2 
\end{equation}
\begin{figure}[h!]
\centering
\includegraphics[width=0.4\textwidth]{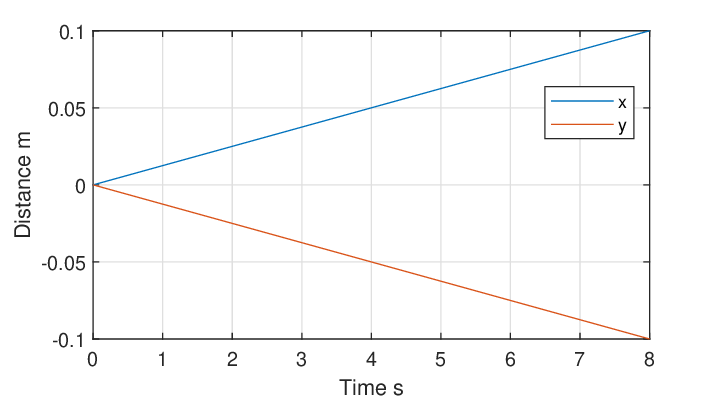}
\includegraphics[width=0.4\textwidth]{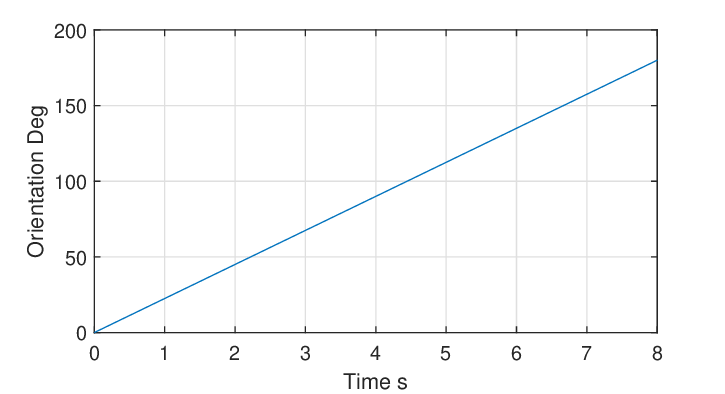}\\
\includegraphics[width=0.4\textwidth]{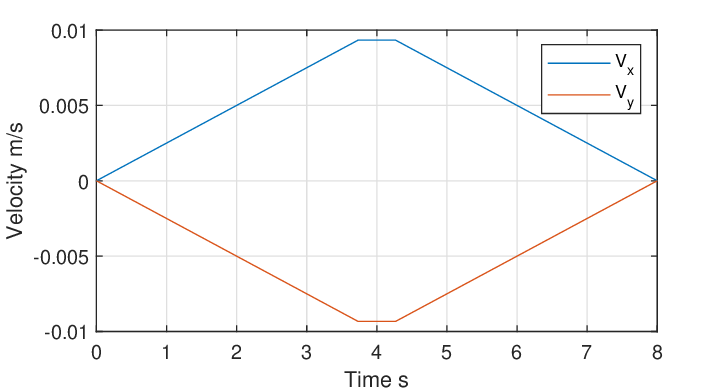}
\includegraphics[width=0.4\textwidth]{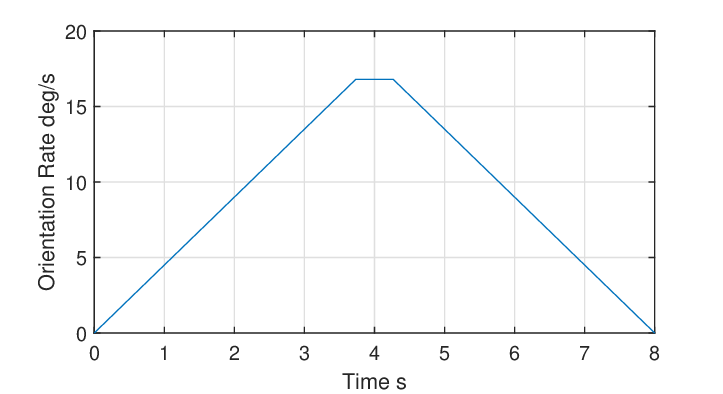}\\
\caption{Guess structure for the fixed end state and fixed end time two point boundary value problem}
\label{guess}
\end{figure}

\begin{figure}[b!]
\centering
\includegraphics[width=\textwidth]{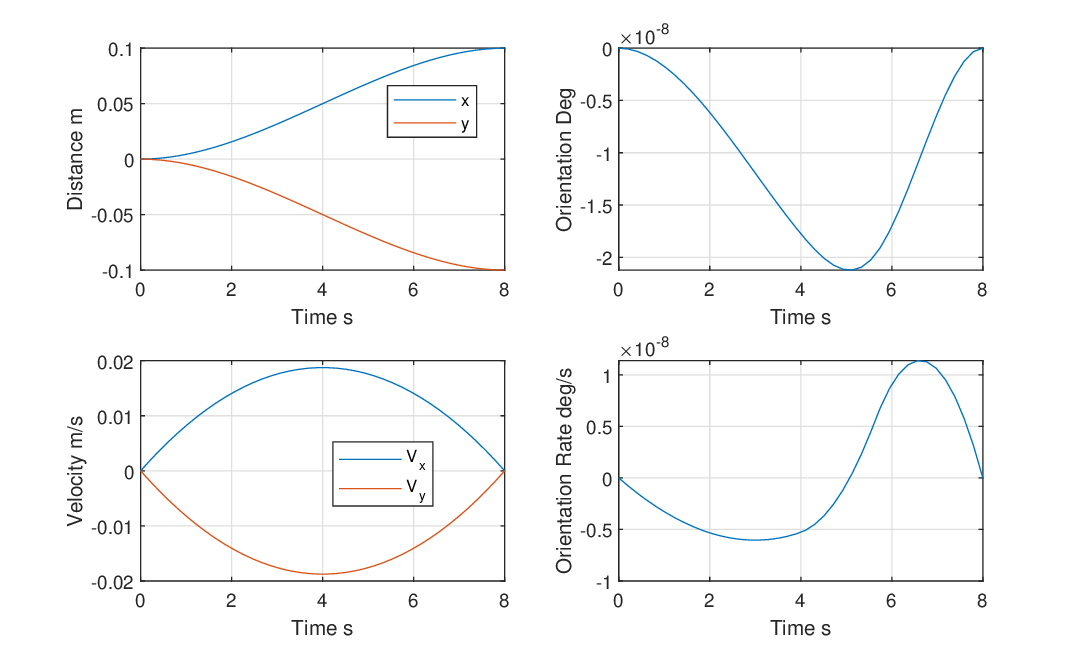}\\
\caption{State history for the fixed end state and fixed end time two point boundary value problem with unconstrained inputs}
\label{states_un}
\end{figure}

\subsection{Formulation and Solution of Two Point Boundary Value Problem}
Euler-Lagrange equations\cite{kirk2012optimal} can be utilized to find governing dynamical equations for states and co-states from the revised Hamiltonian of Equation \ref{H_revised}. This results in six differential equations for states and six differential equations for co-states, systems of equations. For a fixed finite time of the motion $t_f$, twelve boundary conditions are required to propagate the dynamical system from the initial state to the final state optimally. The twelve boundary conditions can be enforced as the six initial states and six desired states at the finite final time $t_f$. Consequently, a total of twelve differential equations along with the same number of boundary conditions form a Two-Point Boundary Value Problem(TPBVP) and are solved using fourth-order collocation-based solver \textit{bvp4c} from MATLAB{\textregistered}. One of the key parameters that greatly affects the computational efficiency, even the possibility of finding a valid solution, is the provided initial guess structure for states and co-states. Poorly formed initial guess structure often results into a singular Jacobin, a vital term to propagate states from initial to desired conditions satisfying the dynamical governing equations. An educated guess structure as shown in Figure \ref{guess}, is employed for the presented analysis. The guess structure assumes that the position and orientation vary linearly with time and that the respective velocities are dynamically consistent with their position/orientation counterparts. For the co-states, a small value of $1e-4$ is chosen for the states that remain constant throughout the motion. For the remaining variables, linear variation with the time having a slope of $1e-6$ has been employed. This guess behavior is mainly driven by the observation of Equations \ref{optimal_u} and governing equations of co-states. 

\begin{figure}[b!]
\centering
\begin{subfigure}{0.49\textwidth}
    \includegraphics[width=\textwidth]{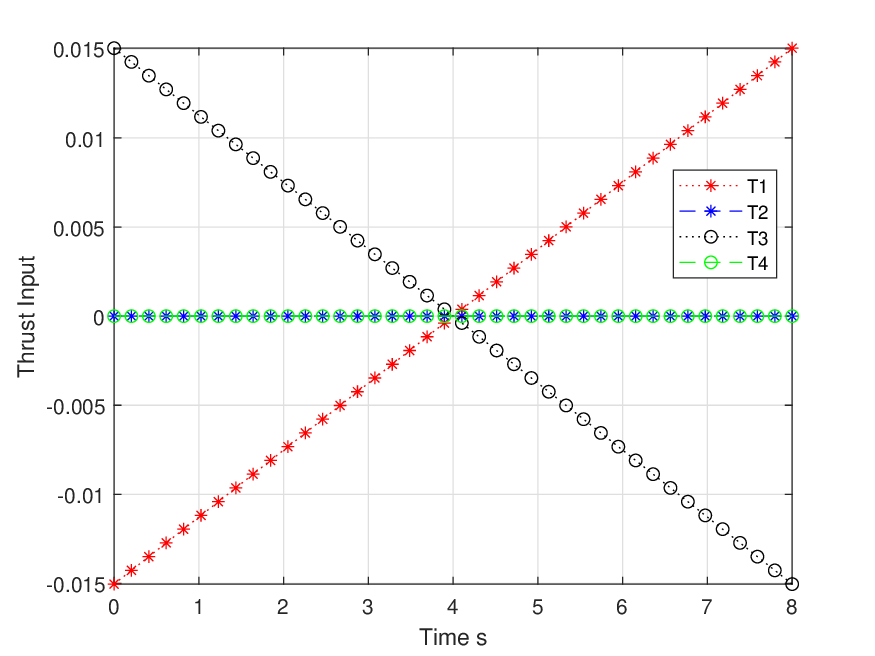}
    \caption{}
    \label{output_un}
\end{subfigure}
\hfill
\begin{subfigure}{0.49\textwidth}
    \includegraphics[width=\textwidth]{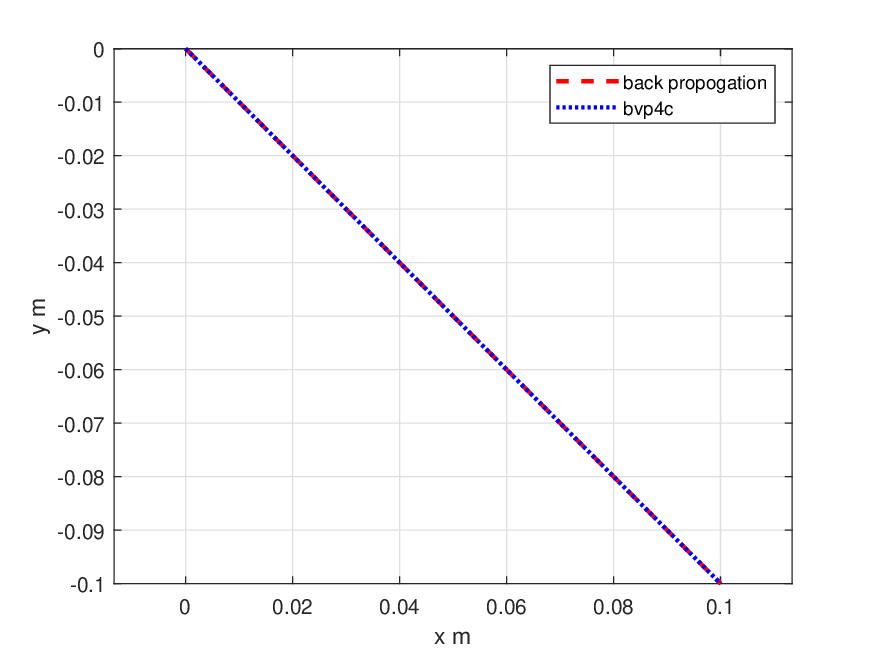}
    \caption{}
    \label{output_un_b}
\end{subfigure}
\caption{Optimal unconstrained inputs and comparison of back propagated trajectory with solution of the fixed end state and fixed end time two point boundary value problem}
\label{fig:figures}
\end{figure}

With the provided guess structure of Figure \ref{guess}, an optimal trajectory that drives initial states [0,0,0,0,0,0] to desired final state [0.1,-0.1,0,0,0,0] in fixed final time $t_f$ is presented in the Figure \ref{states_un}. While the states appear to be honoring the required boundary conditions and the dynamical governing equations, the optimal control history presented in Figure \ref{output_un} reveals a glaring issue that makes such a trajectory infeasible. As discussed in the earlier section, the thrusters are assumed to be unidirectional, and the assumption is constant with, if not all, majority of the practical cold-gas thrusters. However, the current formulation does not prevent the thrust inputs from becoming negative. Hence the optimal control inputs shown in Figure \ref{output_un} are not feasible. On the contrary, as presented in Figure \ref{output_un_b}, if the states are propagated using the given control inputs and enforcing the governing equations, the resulting trajectory has a close match with the solution of TPBVP. The optimal control inputs of Figure \ref{output_un} are infeasible, they do not violate any of the dynamical governing equations.  

\subsection{Incorporation of Constrained Inputs}
For most of the practical applications, the unidirectional thrusters provide thrust in the range of $0\leq T \leq T_{max}$. These constraints have to be incorporated in the TPBVP formulation to compute viable optimal control inputs. This can be achieved by employing Pontryagin's maximum principle\cite{kirk2012optimal}, which states that for any input to be extremal, the Hamiltonian with the extremal input has to be less than any other inputs. Equation \ref{PMP} has to hold for all admissible inputs $0\leq u \leq T_{max}$ and during the entire duration of the motion $t\in[0,t_f]$. The necessary conditions for optimality ensure that the Equation \ref{PMP} is satisfied for the entire duration of the motion, provided the input is unconstrained. However, in the case of constrained input, an explicit analysis is required to compute extremal input that will satisfy Equation \ref{PMP}.      
\begin{align}
H(x^*,u^*,\lambda^*) \leq H(x^*,u,\lambda^*) \nonumber \\
\dot{x}^* = \frac{\partial H}{\partial \lambda} \quad \dot{\lambda} = -\frac{\partial H}{\partial x}
\label{PMP}
\end{align}
\begin{equation}\label{T1}
0 \leq \frac{1}{\sqrt{2}m}(\lambda_4-\lambda_5)-\frac{d}{I_{zz}}\lambda_6 \leq T_{max}
\end{equation}
\begin{equation}\label{H_T1}
H = \frac{1}{2}(T_1^2)-T_1\left(\frac{\left(\lambda_4-\lambda_6\right)}{m\sqrt{2}}-\lambda_6\frac{d}{I_{zz}}\right)
\end{equation}

Without loss of generality, the rationale for selecting extremal constrained input is presented via Equations \ref{T1} and \ref{H_T1}. From the earlier analysis, Equation \ref{optimal_u} gives the extremal control input $T_1$. If, at a particular instance, the required extremal input is within the feasible limits, then the control input $T_1$ is replaced by the extremal $T_1^*$ in Equation \ref{H}. From Equation \ref{H_T1}, which isolates terms containing $T_1$ only from the total Hamiltonian Equation \ref{H}, it can be proven that for any demanded $T_1^*<0,T_1=0$ minimizes Equation \ref{H_T1}. Similarly, for any demanded $T_1^*>T_{max},T_1=T_{max}$ minimizes Equation \ref{H_T1}. Such analysis can be extended to the remaining control inputs, resulting into a general hypothesis as $T=0$ and $T=T_{max}$ minimizes the overall cost function in the case of $T^*<0$ or $T^*>T_{max}$, respectively. As a consequence, the revised Hamiltonian given by Equation \ref{H_revised} no longer holds for instances of constrained inputs. Such a situation can be handled by following the overall process summarized in the Algorithm \ref{alg:PMP}.  

\begin{algorithm}
\caption{Incorporation of Constrained input}\label{alg:PMP}
\begin{algorithmic}[1]
\State Assume At each step, compute values of control inputs assuming unconstrained optimality conditions given by Equations \ref{optimal_u}
\State Observe for the violation of the control input constraints and replace the contribution of the saturated input accordingly.
\State Compute the state and co-state derivatives considering the constrained and unconstrained control inputs. 
\State Propagate the state and co-state by following the same procedure of Step 1-3 in during each time step. 
\end{algorithmic}
\end{algorithm}
\begin{figure}[h]
\centering
\includegraphics[width=\textwidth]{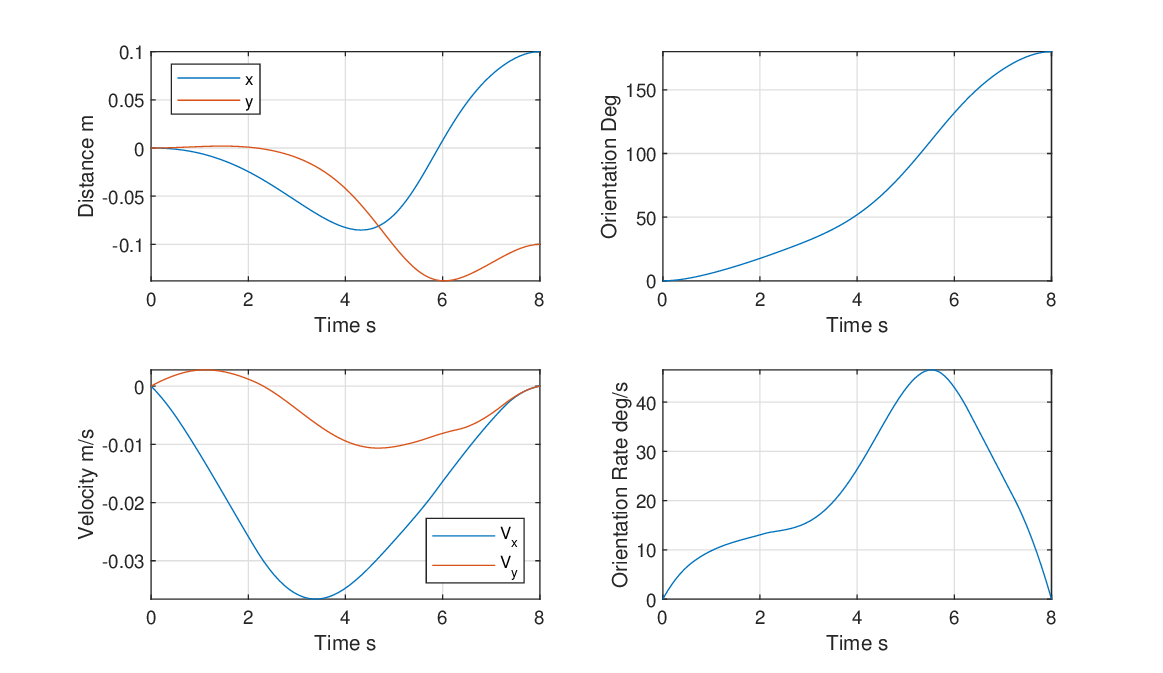}\\
\caption{State history for the fixed end state and fixed end time two point boundary value problem}
\label{optimal_states}
\end{figure}

\begin{figure}[h!]
\centering
\begin{subfigure}{0.45\textwidth}
    \includegraphics[width=\textwidth]{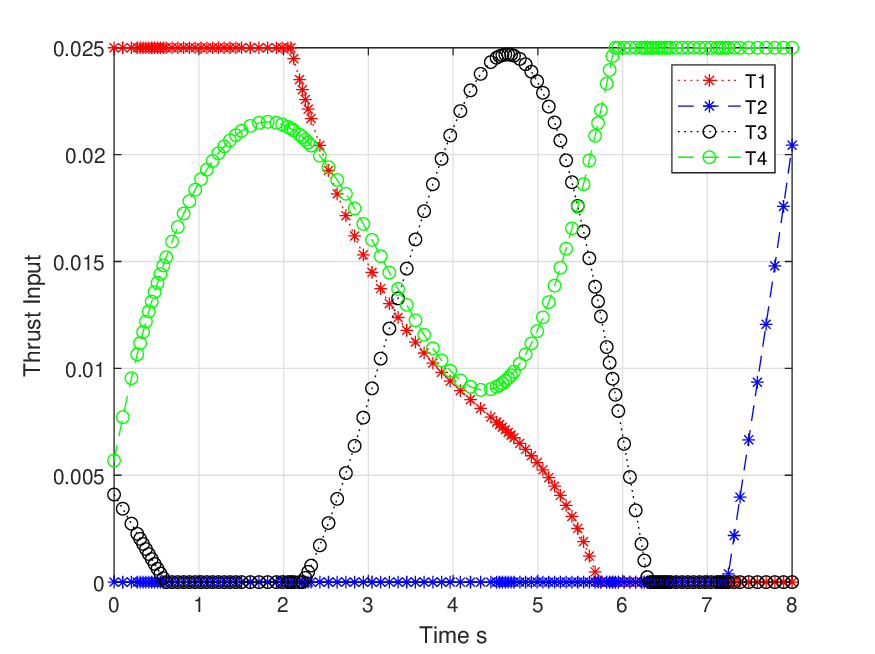}
    \caption{}
    \label{output}
\end{subfigure}
\begin{subfigure}{0.45\textwidth}
    \includegraphics[width=\textwidth]{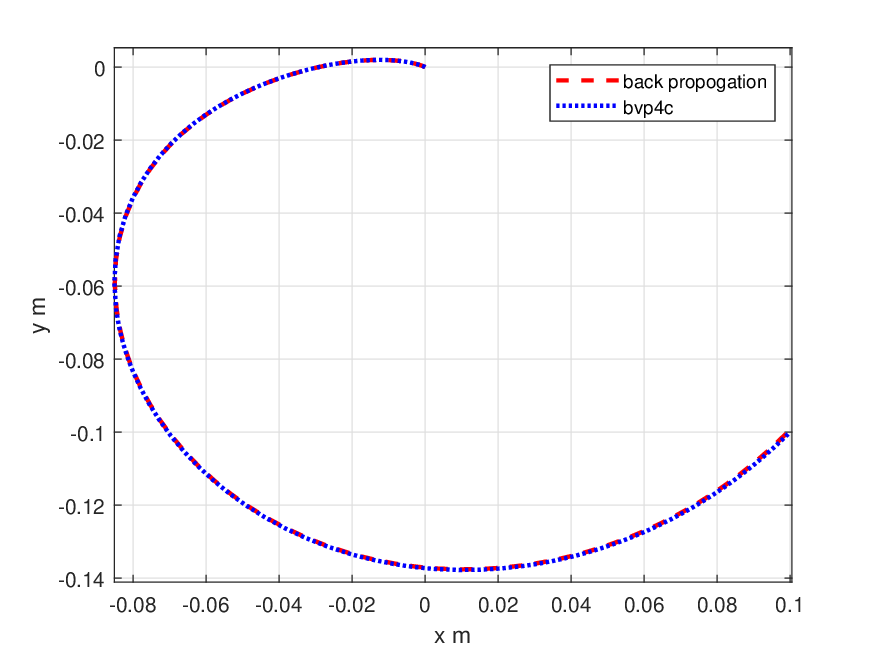}
    \caption{}
    \label{output_b}
\end{subfigure}
\caption{Optimal inputs and comparison of back propagated trajectory with solution of the fixed end state and fixed end time two point boundary value problem}
\label{fig:figures}
\end{figure}

With the provided guess structure of Figure \ref{guess} and an additional check depicted in Algorithm \ref{alg:PMP}, an optimal trajectory that drives initial states [0,0,0,0,0,0] to desired final state [0.1,-0.1,0,0,0,0] in fixed final time $t_f$ is presented in the Figure \ref{optimal_states}. For the current analysis $T_{max}$ value of $25$ mN is selected, which is consistent with the commercially available of-the-shelf micro thruster systems\cite{vacco}. The states appears to be honoring the required boundary conditions and also the dynamical governing equations, the optimal control history presented in Figure \ref{output} testifies the effectiveness of the Algorithm \ref{alg:PMP}. The optimal control inputs for extermal trajectory remains withing the  range throughout the motion. Hence the optimal control inputs shown in Figure \ref{output} can be easily employed for practical applications. In addition, as presented in Figure \ref{output_b}, if the states are propagated using the given control inputs and enforcing the governing equations, the resulted trajectory has a close match with the solution of TPBVP, proving that optimal control inputs of Figure \ref{output} are feasible and consistent with the dynamical governing equations. It is also important to note that the overall trajectory shown in Figure \ref{output_b} for the case of constrained inputs is not identical to the case of unconstrained inputs, given in Figure \ref{output_un_b}. Both trajectories only match in the case where none of the control inputs crosses the enforced limits.

\section{Design and Fabrication of Hardware Prototype}
With the theoretical foundation of overall TPODS behavior laid out via comprehensive analysis of the system dynamics, a detailed system design exercise has been conducted next. The overall system architecture is broadly divided into vision, thrusters, and structural subsystems. Each of these are subjected to iterative comparisons, selection, and experimentation cycles to prove concept of operation, concluding with a thorough integration analysis to ensure the inter-compatibility pertaining to mechanical, electrical and communication interfaces of various subsystems. The following subsections summarize the approach and outcome of these processes for individual subsystems. 

\subsection{Thruster Subsystem}
\begin{figure}[t!]
\centering
\rotatebox[origin=c]{90}{\includegraphics[width=0.45\textwidth]{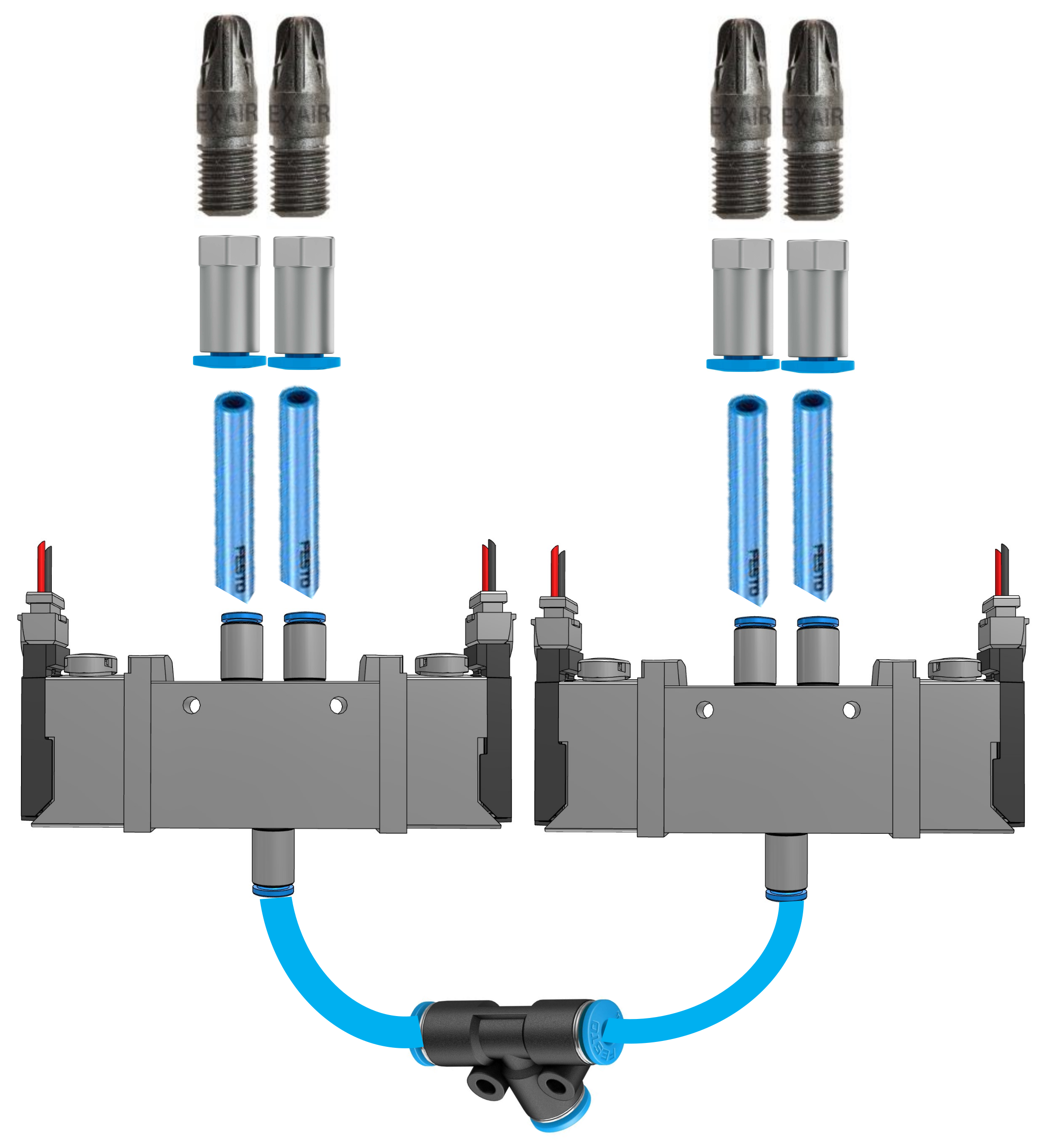}}
\includegraphics[width=0.45\textwidth]{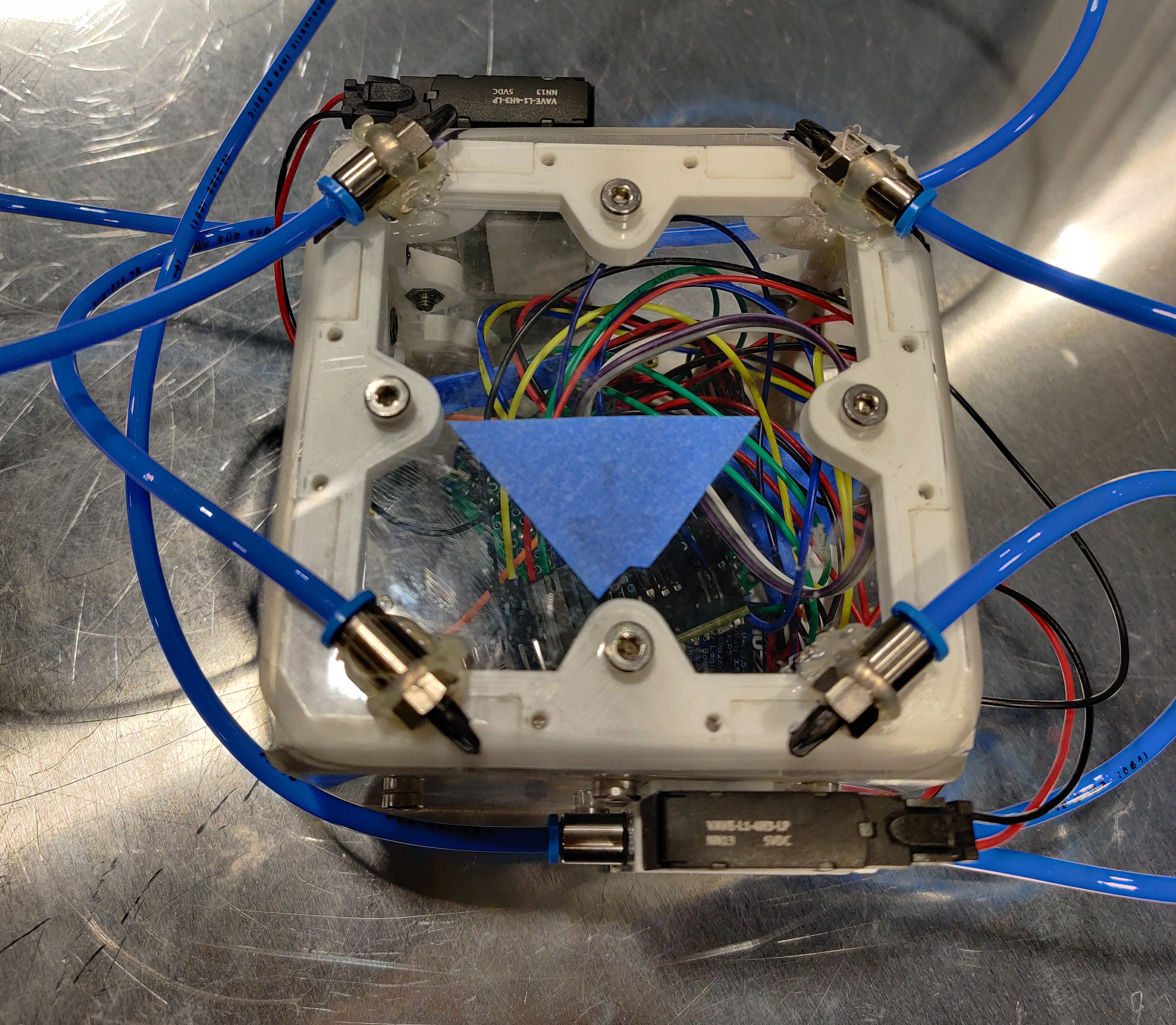}
\caption{Pneumatic Thruster Subsystem of the TPODS satellite module: courtesy of Festo Corporation and EXAIR Corporation}
\label{thruster_system}
\end{figure}

 It is vital to select the locomotion system for the tabletop experiment to resemble the overall behavior of the satellite module with cold gas thrusters. Any RPOD algorithms validated with an equivalent thruster system can be aptly translated from the laboratory environment to the actual mission design. Consequently, compressed air is selected for the thruster system during tabletop experiments. The primary components of the thruster systems are shown in Figure \ref{thruster_system}, which include solenoid valves to rapidly provide the airflow, subsequent tubing and couplers to distribute the air, and nozzles which streamlines the exiting flow to ensure a predictable and smooth reaction force.

\subsubsection{Selection of Valves :}
Most commercially available valves that can be remotely operated have some actuation mechanism attached to the valve stem. One of the primary requirements for the attitude control of satellite modules is rapid switching operation. The electro-mechanical or solenoid-operated valves have fast switching times, sacrificing tight seal at higher pressure differentials. However, as discussed in the following section, for the tabletop experiment the required force, and hence airflow requirement drives operating pressure of the thruster system in a range that the solenoid valves can handle. Consequently, solenoid valves are used to control the airflow rapidly, and in-turn reaction forces on the satellite module. For this particular application, FESTO{\textregistered} \textit{VUVG-L10-T32C-AH-Q4-4H3L-W1} \cite{festo} solenoid valves are used due to their compact form factor, less weight, and fast switching time. It is important to note that the 5V DC operating configuration is selected to eliminate the additional need for high voltage power system for the satellite module.  

\subsubsection{Selection of Nozzles :}
Although the solenoid valves provide rapid and controlled airflow as and when required, directly exposing the exhaust port to the atmosphere results in a rapid and turbulent air stream. This necessitates frequent air compressor operation and poor control of the TPODS module. To tackle this issue, the outgoing airflow is passed to special-purpose nozzles before exposing it to the atmosphere. This final step enables a streamlined airflow, resulting in lower air consumption. The EXAIR{\textregistered} \textit{Atto Super Air Nozzle™} are selected for this application due to its ability to enhance airflow using Coandă effect and to deliver precise jet-stream. It is also worth noting that since the operating environment for these nozzles is tightly controlled, PEEK plastic material is selected. 

\subsubsection{Thrust Modulation :}
Although the combination of solenoid valves and nozzles provides a streamlined constant airflow and consequently a deterministic constant reaction force, the lack of ability to control the reaction force with finite discrete steps results in oscillatory performance of closed-loop position and orientation controller. Such oscillations are prominent around the operating point, as the maximum possible force must correct any small deviation from respective thrusters. However, this can be overcome by utilizing the fast-switching solenoid valves and inertial properties of the TPODS module.

\begin{figure}[b!]
\centering
\begin{subfigure}{0.49\textwidth}
    \includegraphics[width=\textwidth]{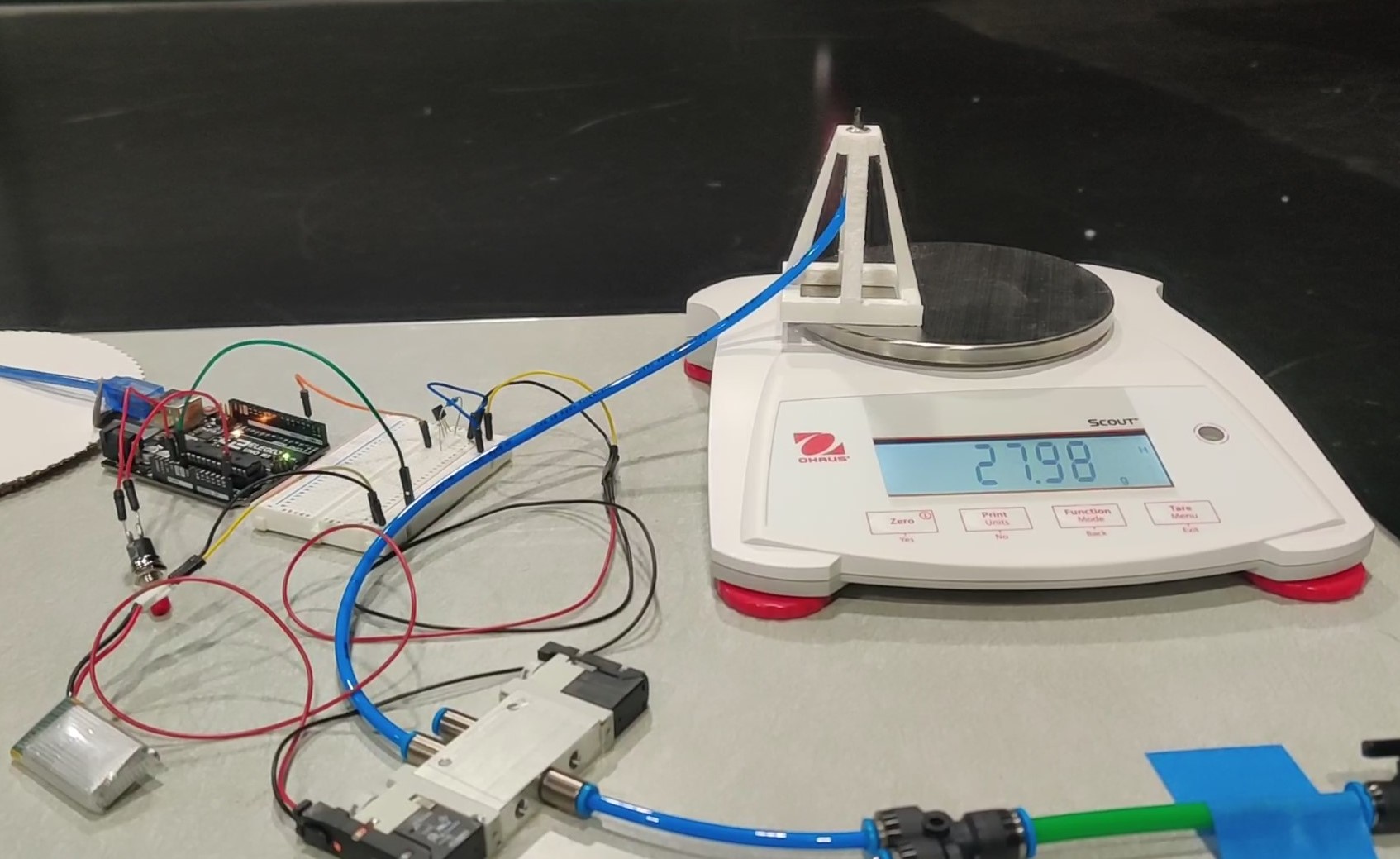}
    \caption{}
    \label{thrust_modulation_setup}
\end{subfigure}
\hfill
\begin{subfigure}{0.49\textwidth}
    \includegraphics[width=\textwidth]{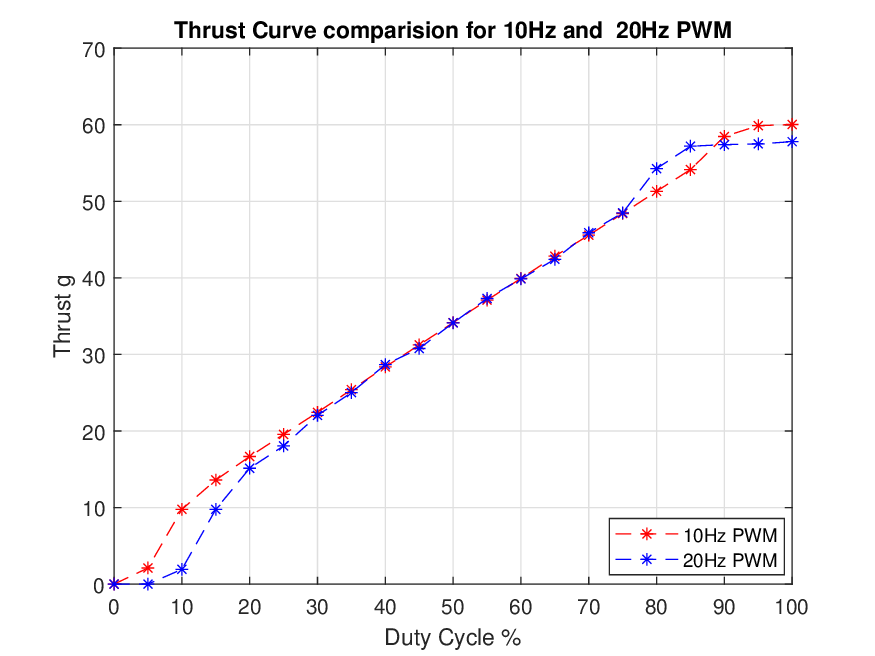}
    \caption{}
    \label{thrust_modulation}
\end{subfigure}
\caption{Experimental Setup to measure modulated thrust and results from the modulation experiment}
\label{fig:figures}
\end{figure}

It has been shown that popular power modulation techniques like Pulse Width Modulation (PWM) can also be applied to pneumatic systems with varying effectiveness based on the ability of various pneumatic components in the system to handle fast switching signals\cite{PWM}. Similar approach is employed to achieve thrust modulation in finite discrete steps. FESTO{\textregistered} \textit{VUVG-L10} require 6 ms to transition from off to on state and 15 ms to transition from on to off state\cite{festo}. Hence, it is ineffective to use higher frequency PWM signals. Figure \ref{thrust_modulation_setup} presents the setup that is used to characterize the effectiveness of the thrust modulation scheme. The nozzle was mounted on  a 3D printed stand rested on a weighing scale calibrated to measure the weight in the range of $0-230g$ with a resolution of  $0.01g$. The nozzle is further connected to a solenoid valve which is being actuated by a 10Hz or 20Hz PWM signals. Starting from 0, the duty cycle of the PWM signal was increased by $5\%$ every 10 seconds and the subsequent weight values are recorded for further analysis. The result from the modulation experiment presented in Figure \ref{thrust_modulation_setup}, revels that the modulation technique is highly effective in providing definite levels of reaction forces based on the duty cycle. Comparison of two thrust plots of 10Hz and 20Hz PWM frequencies in Figure \ref{thrust_modulation} also revel the fact that 20Hz PWM is ineffective near the limits of duty cycle values. This behavior is caused by the 6ms/15ms On/Off time of the valve, as the valve does not get sufficient time to open or close fully while operating within these regions and the thrust output remains unchanged. Consequently, for the closed-loop control of position and orientation of the TPODS module, 10Hz PWM of varying duty cycle is used.

\subsection{Structural Design}
\begin{figure}[h!]
\centering
\begin{subfigure}{0.43\textwidth}
    \includegraphics[width=\textwidth]{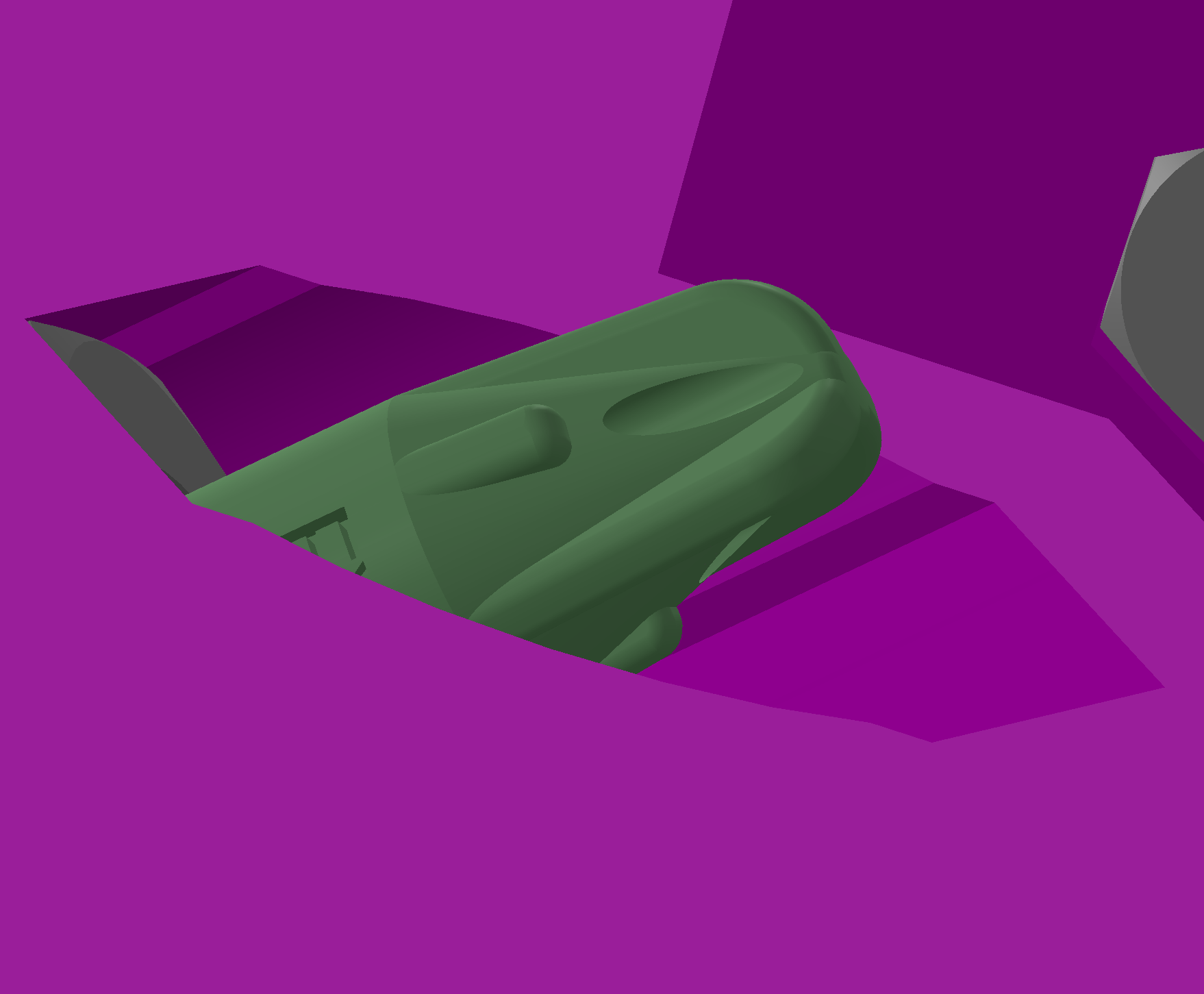}
    \caption{}
    \label{nozzle}
\end{subfigure}
\hfill
\begin{subfigure}{0.44\textwidth}
    \includegraphics[width=\textwidth]{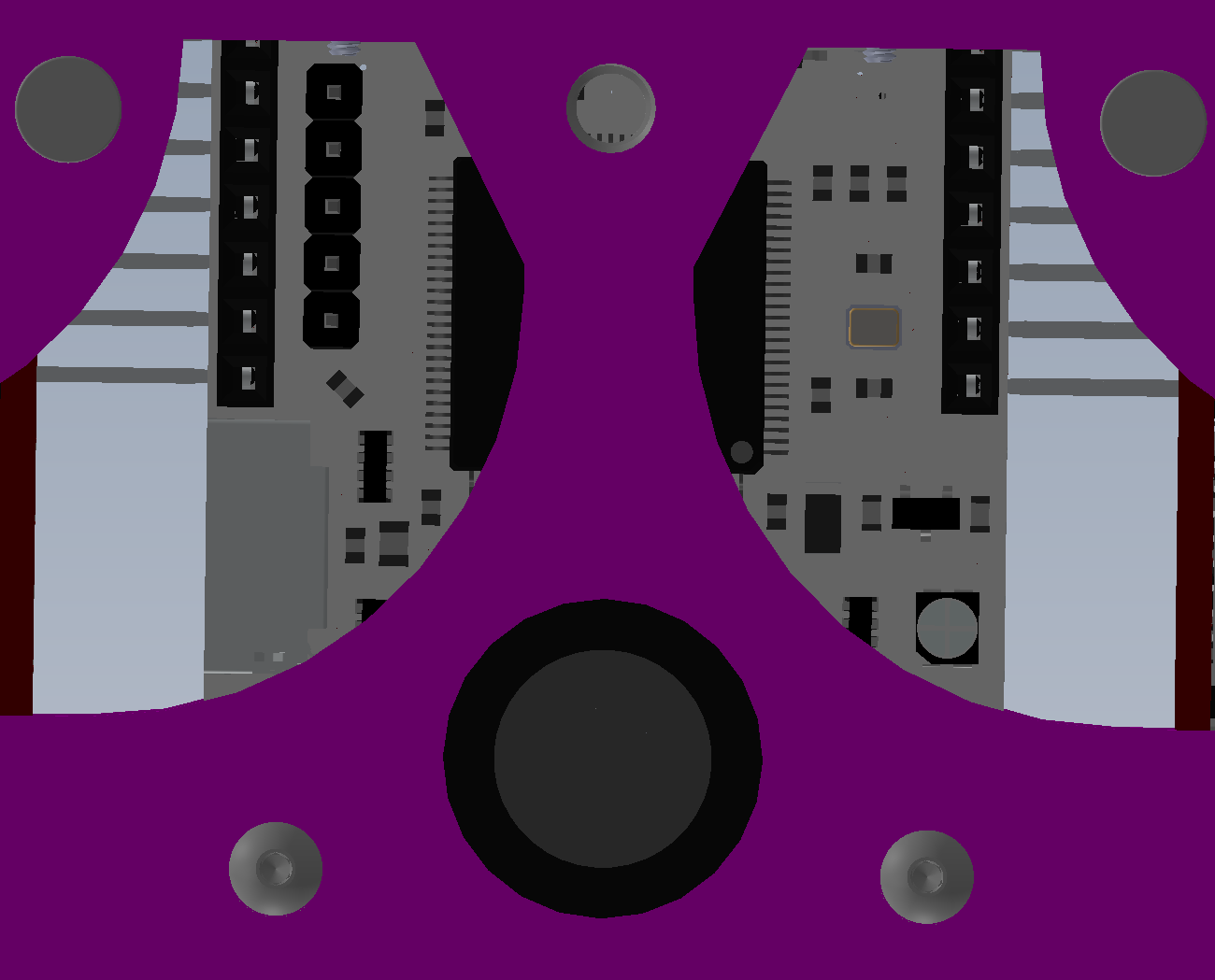}
    \caption{}
    \label{camera}
\end{subfigure}
\caption{Mounting for thrust nozzles and Camera modules}
\label{fig:figures}
\end{figure}

The design process for the structural elements of the TPODS module was initiated by identifying primary needs. First, each TPODS module must be relatively light, with a maximum weight of 5 lbs. Each TPODS structure must also fit within approximate form factor of 1U CubeSat. The functional requirements consist of cavities on each face for led lights, incorporation of thruster nozzles directly to the main frame for efficient and deterministic transfer of reaction forces, sturdy and precise mounting of camera modules on each face, mechanism to enable near-frictionless motion on a planar surface, an attachment mechanism to connect with a rocket body of another TPODS modules and mounting space for relevant electronic and power components. It was also decided to leverage expertise in T-slotted frame structures, acquired by multiple successful design projects at the LASR laboratory, for the initial iteration of the module's main frame\cite{LASR_HAZEL}. However, a comprehensive mechanical size study revealed that such an approach would not be ideal for this application due to the requirement of a compact footprint. With the T-slotted frame structure ruled out, an early skeleton structure was designed such that it could be 3D printed flat and later can be assembled to form a cube. This initial prototype had 1/8$"$ thick walls and magnets embedded into the exterior shell to enable connection of the module with a rocket body or another module. 

The initial skeleton design was 3D printed for fit and finish checks and revealed a few vital issues. Primarily, the 1/8$"$ thick walls were not robust enough to provide structural integrity during and after the interconnection of various faces. To rectify this, the wall thickness was increased to 1/4$"$. Moreover, the location of interconnections was moved for increased firmness. Additionally, cylindrical magnets replaced the cubic magnets, and would protrude from the skeleton through the shell. The camera modules for this iteration were connected directly to the shell via
standoffs. This was the first iteration to be entirely constructed as two full-scale models were manufactured, with one containing electronics and the other serving as a more realistic physical model with spherical ball transfers. With this iteration, the nozzles were retrofitted to the top outer shell and solenoid valves with the side faces, with a makeshift arrangement to check the validity of the locomotion using compressed air thrusters. The final iteration consists of minor improvements to the earlier version while incorporating a thruster system inside the module. The skeleton structure was revised on two sides to contain the nozzles, positioned at 45-degree angles to the center as shown in Figure \ref{nozzle}. Additionally, all sides were enhanced to allow for more stable mounting of camera modules as shown in Figure \ref{camera}. Each camera module threads through the 3D-printed skeleton, connecting directly to the structure via standoffs. Finally, the bottom skeleton side was modified to better house the ball transfers. 

\begin{figure}[h!]
\centering
\begin{subfigure}{0.22\textwidth}
\centering
    \includegraphics[width=\textwidth]{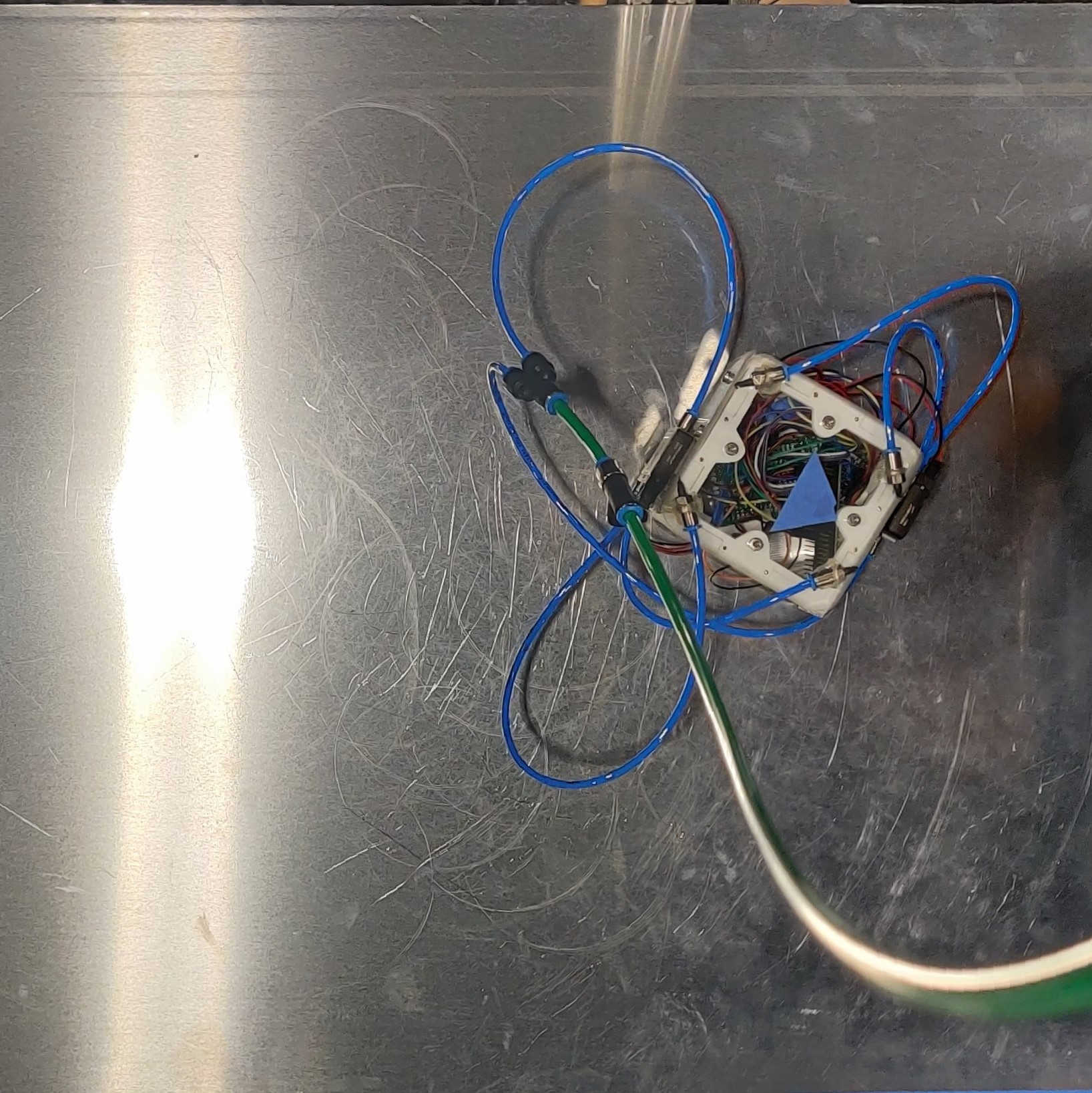}
    \caption{}
    \label{tabletop}
\end{subfigure}
\hfill
\begin{subfigure}{0.44\textwidth}
\centering
    \includegraphics[width=\textwidth]{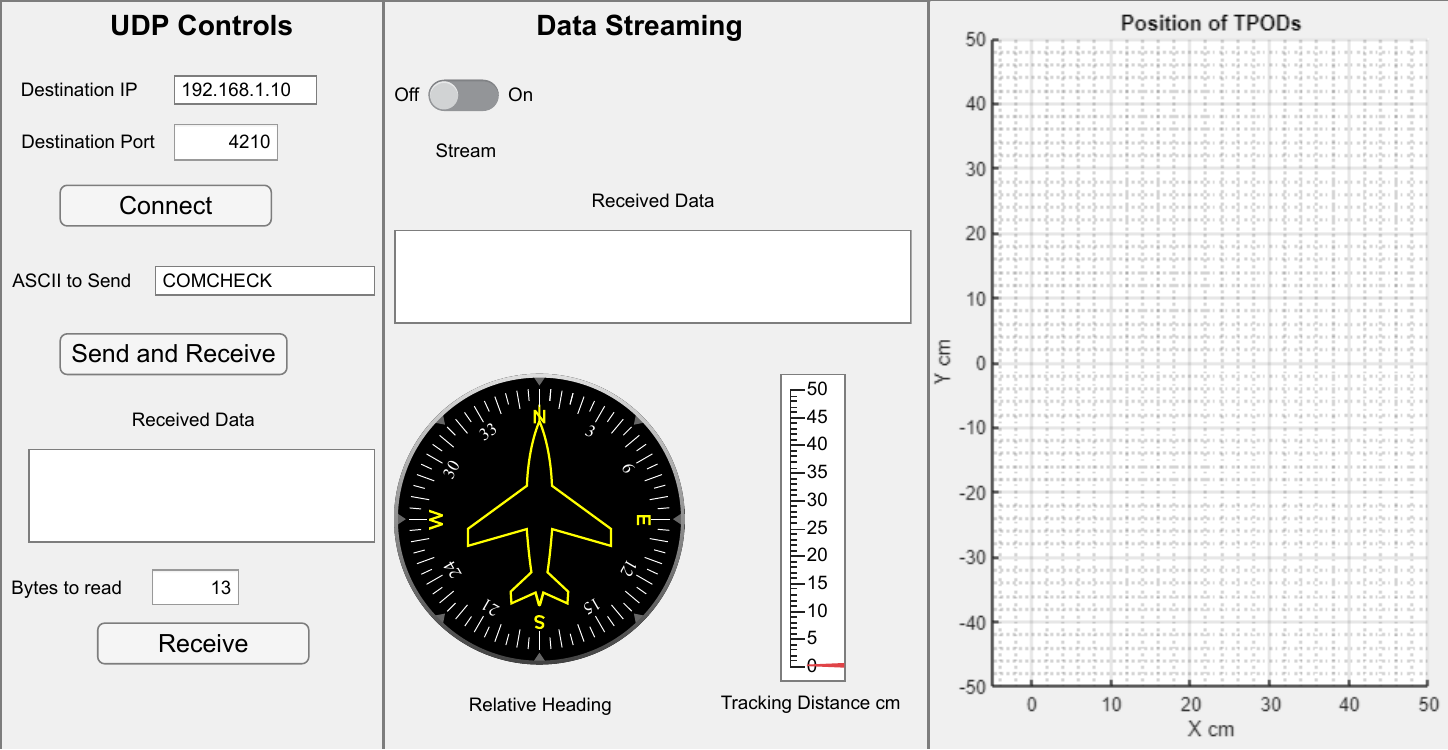}
    \caption{}
    \label{tabletop_GUI}
\end{subfigure}
\hfill
\begin{subfigure}{0.32\textwidth}
    \includegraphics[width=\textwidth]{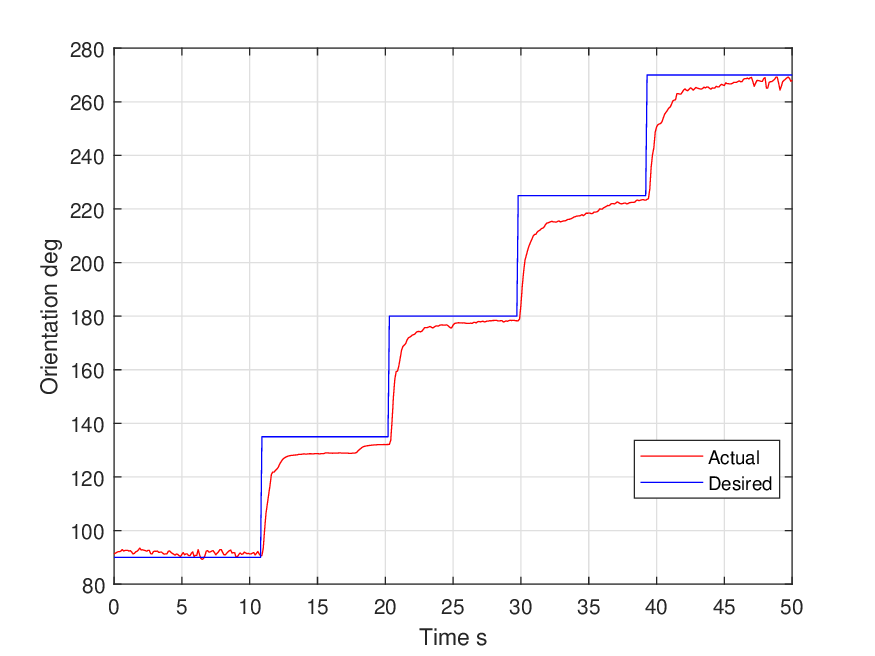}
    \caption{}
    \label{yaw_cl}
\end{subfigure}
\caption{Tabletop experimental setup, GUI for real-time telemetry and tracking, and Desired and Actual orientation during a validation experiment}
\label{fig:figures}
\end{figure}

\section{Tabletop Experiments}
\subsection{Experimental Setup}
The overall goal of the tabletop experiment is to validate the RPOD strategies developed for specific applications of TPODS modules. The satellite module consists of a GY-80 IMU sensor and OpenMV H7 R1 camera modules as the primary sensing elements. The GY-80 IMU sensor is used to independently validate and tune the closed loop orientation controller and is left unused for future explorations while using the OpenMv H7 R1 camera modules for the pose estimation. The teensy 4.1 development board interfaces with the GY-80 IMU sensor and OpenMV H7 R1 camera module and further commands the solenoid valves to generate the required reaction force on the module. However, the GPIO pins of the development board cannot supply enough current to the solenoid valves directly for their nominal operation. Hence, a 2n2222A transistor-based current amplifier is used to boost the current supplied to the solenoid valves. 

The central controller is also connected to the ESP8266 WiFi module via a separate SPI bus. This enables relaying essential information about the system's internal states and parameters to a ground station via WiFi. The ESP8266 WiFi module creates an access point and transfers the data received from the teensy 4.1 board through UDP packets. Figure \ref{tabletop} shows the mast and tethered pneumatic tube supply to the TPODS module. The module is placed on a metal sheet to reduce further the friction between the spherical ball transfers and the surface. Figure \ref{tabletop_GUI} shows the ground control station GUI developed using MATLAB{\textregistered} App. The app connects to the access point created by the WiFi module and requests the streaming of crucial data, which is displayed on the GUI and at the end of the session, gets stored for the offline analysis. 

\subsection{Closed Loop Control of Orientation with IMU measurements}
To validate and tune the cloed-loop attitude controller, the TPODS module was operated using the IMU data as the primary sensing element. The raw data from the onboard accelerometer, gyroscope and magnetometer is accessed by the teensy 4.1 board. The board performs a series of bias rejection, scaling and estimation steps, to compute the orientation data from the raw values of IMU sensors. Once the estimated orientation is calculated using the AHRS subroutine, it is fed to a PI-NZSP-SDR controller to compute the required reaction force in order to achieve the desired orientation. Figure \ref{yaw_cl} testifies the effectiveness of the AHRS, state feedback controller, and the thrust modulation.     

\section{Conclusion}
A comprehensive model-based system design has been executed to realize the conceptual design into a working prototype. Iterative CAD modeling and fabrication using 3D printing enabled rapid prototyping, assessment for areas of improvement, and subsequent revisions. A systematic sensor characterization, noise and bias correction, extensive refinements of estimator parameter, and a compute board capable of running the estimation at a fixed time interval resulted in accurate orientation information. Ample analysis of the dynamical behavior of the TPODS module significantly aided the tuning process of the closed-loop position and orientation controller. Finally, the vigilant architecture design process ensured the seamless integration of numerous electronic and mechanical components. The addition of the ground control station GUI provided vital real-time insight into the current state of various sub-modules, significantly reducing the time and effort required to debug miscellaneous deviations and errors

\section{Extension and Future Work}
A near-term extension of this work includes revamping the ground station and pneumatic supply header to support the simultaneous motion and monitoring of multiple TPODS modules. Additional guidance and navigation strategies can be analyzed for RPOD operations of TPODS modules, followed by the experimental validation of the effectiveness of the theoretical formulations. In the long term, the tabletop system can be modified to allow an experimental analysis of multi-agent-based task allocation problems. A similar process can be followed to devise and realize a true 6-DOF testbed to emulate satellite motion in the microgravity environment.

\section{Acknowledgements}
This work was supported by the Air Force Office of Scientific Research (AFOSR), as a part of  the SURI on OSAM project "Breaking the Launch Once Use Once Paradigm" (Grant No: FA9550-22-1-0093). Program monitors, Dr. Frederick Leve of AFOSR, and Dr. Jason Guarnieri of AFRL are gratefully acknowledged for their watchful guidance. Prof. Howie Choset of CMU, Mr. Andy Kwas of Northrop Grumman Space Systems and Prof. Rafael Fierro of UNM are acknowledged for their motivation, technical support, and discussions.

\bibliographystyle{AAS_publication}   
\bibliography{./bib/refsm-astro}   
\end{document}